\documentclass[reprint,amsmath,amssymb,apr,aip,onecolumn]{revtex4-2}
\usepackage{graphicx}
\usepackage{graphics}
\usepackage{algorithm,algorithmic}
\usepackage{mathptmx}
\usepackage{times}
\usepackage{amsmath}
\usepackage{amssymb, bm}
\usepackage{dcolumn}
\usepackage{color}
\usepackage{bm}
\usepackage{units}
\usepackage{booktabs}
\usepackage{multirow}
\usepackage{xcolor}
\usepackage{url}

\usepackage[acronym]{glossaries}
\newacronym{svm}{SVM}{Support Vector Machine}
\newacronym{mac}{MAC}{multiply-accumulate}
\newacronym{mse}{MSE}{Mean Squared Error}
\newacronym{krr}{KRR}{kernel ridge regression}
\newacronym{rkhs}{RKHS}{reproducing kernel Hilbert space}
\newacronym{gpu}{GPU}{Graphic Processing Unit}
\newacronym{mvm}{MVM}{matrix-vector multiplication}
\newacronym{aimc}{AIMC}{analog in-memory computing}
\newacronym{adc}{ADC}{analog to digital converter}
\newacronym{dac}{DAC}{digital to analog converter}
\newacronym{pcm}{PCM}{phase change memory}
\newacronym{snr}{SNR}{signal-to-noise ratio}
\newacronym{dl}{DL}{deep learning}
\newacronym{ml}{ML}{machine learning}
\newacronym{lra}{LRA}{Long Range Arena}
\newacronym{rff}{RFF}{Random Fourier Features}
\newacronym{orf}{ORF}{Orthogonal Random Features}
\newacronym{sorf}{SORF}{Structured Orthogonal Random Features}
\newacronym{rbf}{RBF}{Radial Basis Function}
\newacronym{rnn}{RNN}{Recurrent Neural Network}
\newacronym{cnn}{CNN}{Convolutional Neural Network}
\newacronym{nlp}{NLP}{Natural Language Processing}
\newacronym{flop}{FLOP}{Floating Point Operation}
\newacronym{tops}{TOPS}{Tera Operations Per Second}
\newacronym{tops/w}{TOPS/W}{Tera Operations Per Second Per Watt}
\newacronym{nvm}{NVM}{Non-Volatile Memory}
\newacronym{mha}{MHA}{Multi Head Attention}
\newacronym{gdp}{GDP}{Gradient Descent-based Programming}

\renewcommand{\arraystretch}{1.3}

\begin{document}
\title{Kernel Approximation using Analog In-Memory Computing}

\author{Julian B\"{u}chel}\email{jub@zurich.ibm.com}\affiliation{These authors contributed equally}\affiliation{IBM Research Europe, 8803 R\"{u}schlikon, Switzerland}
\author{Giacomo Camposampiero}\affiliation{These authors contributed equally}\affiliation{IBM Research Europe, 8803 R\"{u}schlikon, Switzerland}
\author{Athanasios Vasilopoulos}\affiliation{IBM Research Europe, 8803 R\"{u}schlikon, Switzerland}
\author{Corey Lammie}\affiliation{IBM Research Europe, 8803 R\"{u}schlikon, Switzerland}
\author{Manuel Le Gallo}\affiliation{IBM Research Europe, 8803 R\"{u}schlikon, Switzerland}
\author{Abbas Rahimi}\affiliation{IBM Research Europe, 8803 R\"{u}schlikon, Switzerland}
\author{Abu Sebastian}\email{ase@zurich.ibm.com}\affiliation{IBM Research Europe, 8803 R\"{u}schlikon, Switzerland}

\date{\today}

\begin{abstract}
Kernel functions are vital ingredients of several machine learning algorithms, but often incur significant memory and computational costs. We introduce an approach to kernel approximation in machine learning algorithms suitable for mixed-signal Analog In-Memory Computing (AIMC) architectures. Analog In-Memory Kernel Approximation addresses the performance bottlenecks of conventional kernel-based methods by executing most operations in approximate kernel methods directly in memory. The IBM HERMES Project Chip, a state-of-the-art phase-change memory based AIMC chip, is utilized for the hardware demonstration of kernel approximation. Experimental results show that our method maintains high accuracy, with less than a 1\% drop in kernel-based ridge classification benchmarks and within 1\% accuracy on the Long Range Arena benchmark for kernelized attention in Transformer neural networks. Compared to traditional digital accelerators, our approach is estimated to deliver superior energy efficiency and lower power consumption. These findings highlight the potential of heterogeneous AIMC architectures to enhance the efficiency and scalability of machine learning applications.
\end{abstract}

\maketitle

\section{Introduction}
Kernel functions\cite{learningkernels, kernelreview} of the form $k(x,y)$, where $x$ and $y$ are $d$-dimensional input vectors, are specialized mathematical constructions that are equivalent to the computation of scalar measures such as similarities or inner products between data points in some designated higher, possibly infinite, dimensional space, but can be evaluated exclusively within the original space. Because kernel functions allow to capture non-linear relationships in the input data while operating in the original space, they can be found in many learning algorithms with a wide range of applications, including classification\cite{bosersvm}, regression\cite{svr}, and dimensionality reduction\cite{kerpca}.

However, algorithms based on kernel functions are often hampered by the inability to intertwine the evaluation of the kernel function with other linear operations. This often incurs high computational cost during inference, as some algorithms require the evaluation of the kernel function $k(x_i,y)$ between the new sample $y$ and every data point $x_i$ in the training data. Furthermore, the space cost also grows quadratically with the number of training samples as some algorithms require to store the Gram matrix $\mathcal{G}$, where $\mathcal{G}_{i,j}=k(x_i,x_j)$ for all the pairs $x_i,x_j$ in the training set. To address this scalability issue, researchers have devised a family of techniques known as kernel approximation methods, which provide an efficient means of approximating kernel computations while maintaining competitive predictive performance and strong mathematical guarantees on their representational power. These techniques seek to strike a balance between computational efficiency and accuracy, making it feasible to use kernel functions in large-scale problems. The essential idea behind kernel approximation is to explicitly map the input vectors $x,y \in \mathbb{R}^d$ to some $D$-dimensional space, where $D > d$, using a mapping function $z : \mathbb{R}^d \mapsto \mathbb{R}^D$ so that 
\begin{equation}
k(x,y) \approx \langle z(x), z(y) \rangle.
\end{equation}
A series of algorithms following this idea have been proposed over the past years\cite{liu2021random}, among which a prominent role is played by algorithms based on random feature sampling such as random Fourier features\cite{rahimirecht} and its enhanced variants\cite{orthogonalrff, rom, fastfood, qmc, ssf, gq, lsrff}.

Kernel approximation based on random features finds a wide range of applications, such as kernel \glspl{svm}\cite{kernelsvm1, kernelsvm2, kernelsvm3, kernelsvm4}, dimensionality reduction\cite{dimred1}, neural networks\cite{NIPS2009_5751ec3e, dnn2, dnn3, dnn4}, and kernel regression\cite{kerreg1, kerreg2}. This technique can be used to approximate a particular family of kernels, called shift-invariant kernels, and allows to approximate $k$ using explicit mappings obtained by sampling weights from probability functions that are specific to that kernel\cite{rahimirecht}.
Although the speed-up offered for large-scale problems is already significant, these approximation algorithms based on random features leverage expensive mappings from the original space ($\mathbb{R}^d$) to the feature space ($\mathbb{R}^D$) for every input sample. When implemented on conventional von Neumann architectures, this mapping operation is costly in terms of energy consumption and latency as the weights need to be constantly fetched from memory. This limitation is commonly referred to as the von Neumann bottleneck.

One way to overcome this bottleneck is \gls{aimc}~\cite{Y2020sebastianNatNano,lanza2022, Y2023mannocciAPLML}, a technique in which certain operations can be directly performed in-memory using analog computation. For example, SRAM-based \gls{aimc}~\cite{sram-aimc} is an instance of \gls{aimc} using volatile memory that is fast and has good endurance, but has poor scalability in terms of capacity due to limitations in the SRAM cell size. Another variant of \gls{aimc} uses \gls{nvm} devices such as Flash~\cite{flash-aimc}, MRAM~\cite{mram-aimc}, \gls{pcm}~\cite{mt-hermes} or ReRAM~\cite{rram-aimc,rram-science}. The advantage of using \gls{nvm} devices is their high density. However, intrinsic device noise, inter-device variability, and limited endurance impact the computational accuracy in applications. One application that does not require very high endurance and high-precision computation is the inference of neural networks, which has been widely studied in prior works~\cite{mt-hermes,ambrogio-nature,rram-science}. Here, the weights of the layers of the neural network are programmed on the \gls{nvm} devices arranged in a crossbar array. When applying voltage pulses to the rows of the crossbar array, current flows through the \gls{nvm} devices and accumulates along the column. The resulting current, which represents a dot product between the input voltages and the \gls{nvm} devices in the column, is converted back to the digital domain using \glspl{adc}. Because the crossbar array has multiple columns, multiple dot products are computed, which is the equivalent to a matrix-vector multiplication. By exploiting Kirchhoff’s current law and Ohm’s law, all the multiply-accumulate operations that take place inside the crossbar are executed in parallel. Therefore, one considers the \gls{mvm} operation as one primitive operation which can be executed in one time step. The duration of this step depends on various factors such as the encoding scheme of the 8-bit inputs to voltage pulses and the \gls{adc} conversion time. For more details, see \citealt{Sebastian2020}. By combining such analog computing with light-weight digital processing units on a heterogeneous architecture, \gls{aimc} can provide $10\times$ to $140\times$ higher energy efficiency~\cite{tvlsi} at competitive throughput compared to modern GPUs~\cite{a100}.

Leveraging this idea, we propose in-memory kernel approximation, which allows to accelerate  kernel approximation methods based on random features on such a heterogeneous architecture. This is -- compared to prior efforts that accelerated the kernel approximation using GPUs~\cite{choromanski2020rethinking, gpu-accelerated-kernel-approx} -- achieved by implementing the compute-intensive mapping from the original to the feature space using \gls{aimc}. By doing in-memory kernel approximation of non-linear kernels, a vast majority of operations associated with the linear projection to a higher dimensional space are accelerated using analog computing. This is followed by executing some element-wise non-linear operations in digital units as part of kernel approximation which is in sharp contrast with previous methods that needed to perform the entire kernel approximation in digital units.

We experimentally validate the concept of in-memory kernel approximation using a state-of-the-art 14-nm CMOS \gls{pcm}-based mixed-signal \gls{aimc} chip referred to as the IBM HERMES Project Chip~\cite{mt-hermes}. This chip comprises 64 cores, each of which has a $256 \times 256$ crossbar, 256 \glspl{dac}, 256 \glspl{adc} and local digital post-processing circuitry (see Methods). Because there exists one \gls{dac}/\gls{adc} per row/column, fully-parallel \glspl{mvm} can be performed in a single computational time step, leading to a peak throughput of 63.1 \gls{tops} and energy efficiency of 9.76 \gls{tops} per Watt. Furthermore, the peak power consumption of 6.5W of the IBM HERMES Project Chip is significantly lower than that of a GPU. For example, the NVIDIA A100 GPU has a peak power consumption of 400W of which 70W are static. 
While analog computing suffers from noise and hence, at present, only allows for low precision computations, several contemporary applications in Machine Learning can be made robust to low precision computations~\cite{mt-hermes}.
We first study a representative pool of kernel approximation techniques and measure the approximation error between the 32-bit floating point (FP-32) software kernel approximation and the in-memory kernel approximation.
Furthermore, we measure the impact of the additional approximation error introduced by the chip on the downstream classification accuracy on a set of well-known machine learning benchmarks, using a simple linear classifier (kernel ridge classification model) on top of the extracted features. Finally, we turn to a more recent application of kernel approximation in linear-complexity Transformer neural networks. We show that in-memory kernel approximation of the Softmax kernel used in kernelized attention yields no degradation in downstream performance compared to the FP-32 software equivalent, while offloading between half and one third of the \glspl{flop} involved in the linear attention computation to \gls{aimc}. This is different from previous approaches to accelerate attention using \gls{aimc}. One previous approach~\cite{hardesa} uses \gls{aimc} to remove tokens that would yield low attention scores. Although this can lead to the removal of up to ~90\% of the tokens, only ~20\% of the \glspl{flop} get offloaded to \gls{aimc} for a sequence length of 4,096. For longer context lengths this percentage further diminishes as the attention computation (which is done in digital) still has quadratic complexity with respect to the sequence length. Other approaches~\cite{micro-attention,attention-in-memory} require frequent re-programming of the crossbar arrays during inference. Our approach requires only a one-time programming of the memory devices, and also makes the attention mechanism faster and more memory-efficient as a result of the linear complexity of both time and space with respect to the sequence length.
Assuming peak throughput and peak power consumption, we estimate that in comparison to an NVIDIA A100 GPU (INT8 precision), accelerating the projection operation in kernel-approximation is up to 6.3$\times$ less energy consuming.

\section{Results}
\subsection{Proposed in-memory computing-based acceleration for kernel approximation methods.}
\begin{figure}[h]
\includegraphics[width=\textwidth]{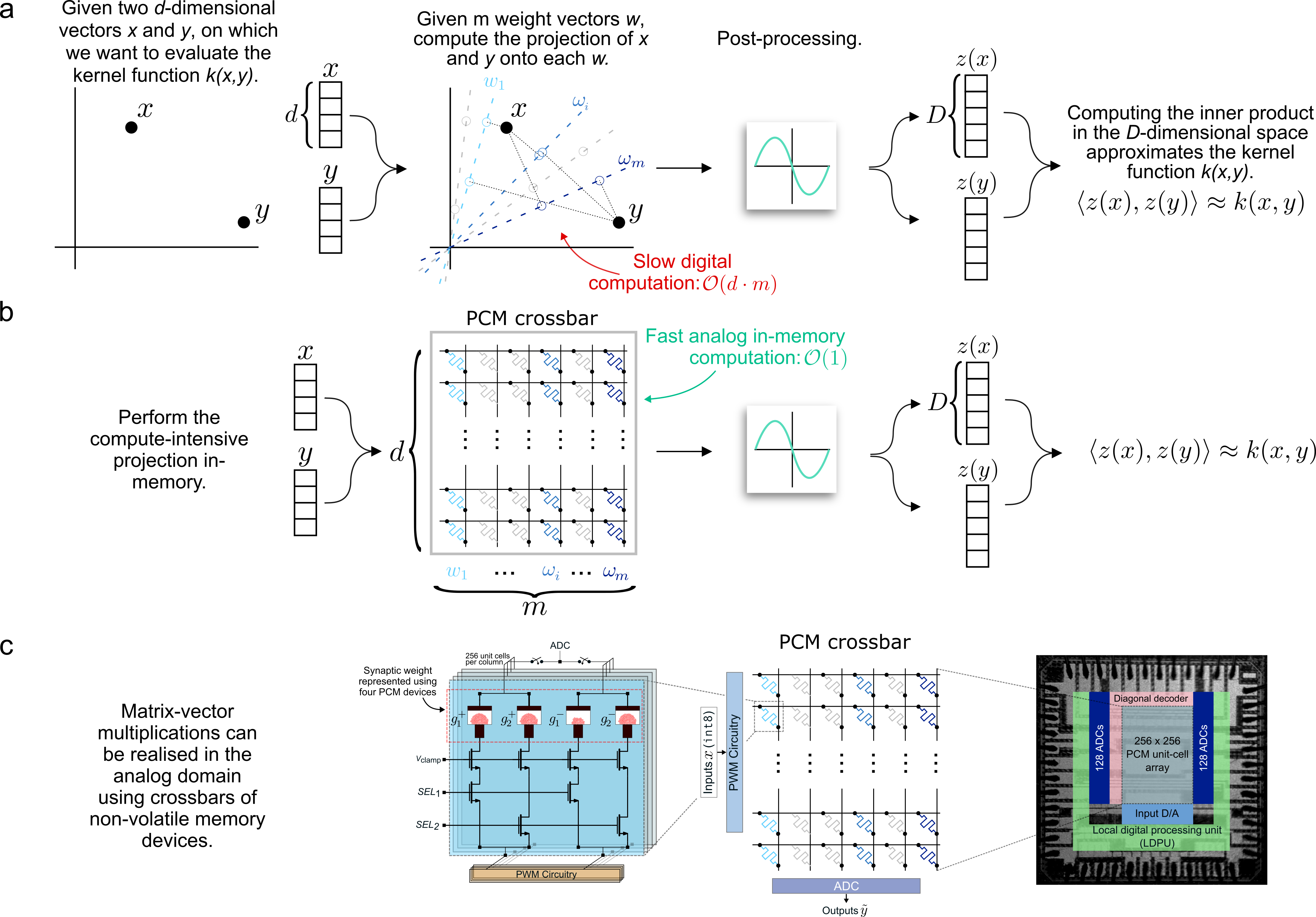}
\caption{\textbf{Proposed in-memory kernel approximation technique.}
\textbf{a} The two vectors $x$ and $y$ for which we want to approximate the kernel function $k$ are projected onto $m$ weight vectors $\omega_i$ that are drawn from the kernel-dependent probability distribution $p(\omega)$. After some element-wise post-processing, one obtains $z(x)$ and $z(y)$ of which the dot product represents the approximated kernel evaluated on $x$ and $y$.
\textbf{b} We program the sampled vectors $\omega_i$ into the columns of a memristive crossbar array and perform the mapping in-memory. The element-wise post-processing and inner product calculation are performed in digital hardware.
\textbf{c} Each weight element of the vectors $\omega$ are represented with four \gls{pcm} devices. These devices are arranged in a crossbar. In order to perform an \gls{mvm}, inputs are quantized to 8 bits and converted to voltage pulses that are then applied to the rows of the crossbar. Current proportional to the weight in the unit cell accumulates across the column, representing a dot product of the input with that column. The accumulated current is converted back to the digital domain using \glspl{adc}. The crossbar array and the peripheral circuits are integrated into one tile.}
\label{fig:intro}
\end{figure}

We now present a more in-depth introduction to the approximation of shift-invariant kernels and describe in-memory kernel approximation in more detail.
In the case of a particular family of kernels, namely shift-invariant kernels, which only depend on the distance between the evaluated data points ($k(x,y) = k(x-y)$), Bochner’s theorem guarantees that the Fourier transform of the kernel $k(\cdot, \cdot)$ is a proper probability distribution. This probability distribution can be used for sampling weight vectors $\omega \in \mathbb{R}^d$ (called random features) that can be used such that, for some real-valued mapping $z_\omega(x): \mathbb{R}^d \mapsto \mathbb{R}^D$, $\mathop{{}\mathbb{E}}[\langle z_\omega(x), z_\omega(y)\rangle ]=k(x,y)$ holds. In other words, the product $\langle z_\omega(x),z_\omega(y)\rangle $ is an estimate of $k(x,y)$. Unfortunately, this estimate has a rather high variance, which is why one typically samples $m$ vectors $\{\omega\}_{i=1}^m$ from the Fourier transform, computes this product $m$ times and averages the result (see Fig.~\ref{fig:intro}a). The resulting mapping $\mathbf{z}$ can be generally expressed as 
\begin{equation} \label{eq:rff}
    \mathbf{z}(x)=\frac{h(x)}{\sqrt{m}}[f_1(\omega_1^Tx),...,f_1(\omega_m^Tx),...,f_l(\omega_1^Tx),...,f_l(\omega_m^Tx)].
\end{equation}
where $h(x)$ is a scaling function, $m$ is the number of random features, and $l$ is the number of functions we use. For example, in the case of \gls{rff}, we have $h(x)=1$, $f_1=cos$ and $f_2=sin$. Since $l=2$ for the \gls{rff} kernel, we see that, because of the concatenation, the resulting mapping dimensionality is $D= l \cdot m = 2 \cdot m = 2 \cdot a \cdot d$, where $a$ is an application-specific factor varying from 2 to 32.

The randomly sampled weight vectors $\omega$, which play an integral part in computing the function $z(\cdot)$ are fixed during inference and do not need to be re-sampled. However, in conventional von Neumann architectures, these $m$ weights still need to be loaded from memory in order to compute the mappings. In-memory kernel approximation avoids this issue by directly computing the mapping in-memory (see Fig.~\ref{fig:intro}b), effectively reducing the number of \gls{flop}s that need to be executed in digital hardware, required to compute $k(x,y)$, from $(8\cdot a\cdot d^2 + 4\cdot l \cdot a \cdot d)$ to $4\cdot l \cdot a \cdot d$. This is achieved by the following procedure.

Initially, each weight vector $\omega$ is mapped onto each column of the crossbar by first concatenating all weight vectors into a single matrix. This matrix is then programmed on the crossbar using a program-and-verify algorithm~\cite{jetcas-buechel}. During inference, incoming FP-32-based input vectors $x$ are first quantized to INT8 using fixed per-crossbar scaling factors, and are then converted to voltage pulses where the width of pulse $i$ is proportional to the $i$-th digital value of $x$. These pulses are then applied to the rows of the crossbar, creating currents that accumulate along each column and feed into one of the 256 \glspl{adc}. The current-controlled oscillator-based \glspl{adc} turn these currents directly into digital counts which are then corrected with an elementwise affine transformation using simple digital circuits in local digital processing units.

To experimentally validate the in-memory kernel approximation, we use the IBM HERMES Project Chip (see Methods for details on the evaluation platform). The mapping weights are programmed onto the analog conductance values of four \gls{pcm} devices organized in a differential configuration with two devices each representing the positive and negative values, respectively\cite{msf}. The mapping operations are subsequently performed by on-chip analog matrix-vector multiplications. 

\subsection{Kernel-based ridge classification.}
First, to study the robustness of kernel approximation techniques to the noise introduced by the analog hardware, we study kernel-based ridge classification on six benchmarks often used to benchmark kernel approximation methods (see Methods). To evaluate the quality of the kernel approximation, we use two metrics: the approximation error between the approximated $\hat{\mathcal{G}} \in \mathbb{R}^{N\times N}$ and original kernel Gram matrix $\mathcal{G} \in \mathbb{R}^{N\times N}$ expressed as
\begin{equation*}
    \text{Approx. Error} = \frac{\|\mathcal{G}-\hat{\mathcal{G}}\|_F}{\|\mathcal{G}\|_F}
\end{equation*}
and the downstream classification accuracy expressed as the percentage of correctly classified inputs. We investigate the performance of two kernels, namely the Gaussian kernel (referred to as \gls{rbf}) and the zeroth-order arc-cosine kernel (referred to as ArcCos0), for which the definition according to Eq.~\eqref{eq:rff} is given in Supplementary Table \ref{tab:kernel_def}. Additionally, we investigate the performance of sampling approaches for $\omega$ that differ from \gls{rff}. In particular, we study the approximation error and downstream performance of \gls{orf}\cite{orthogonalrff}, a technique that proposes to orthogonalize the randomly drawn weight vectors $\omega$  to reduced the required samples. We additionally study \gls{sorf}, which is a direct extension to \gls{orf}, making the generation of the orthogonal matrix cheaper.

In the following, we present a brief overview of kernel-based ridge classification and discuss the advantages of using in-memory kernel approximation over conventional, digital kernel approximation. The predictive function $f: \mathbb{R}^d \mapsto \mathbb{N}$ in kernel-based ridge classification\cite{vovk2013kernel} models is of the form
\begin{equation*}
    f(x) = \text{sgn}\left(\sum_{i=1}^{N}{\alpha_i k(x,x_i)}\right)
\end{equation*}
where $\text{sgn}(\cdot)$ is the sign function, $\alpha_i$ is the $i$-th coefficient of the regression model, $x_i$ is the $i$-th training sample, $N$ is the number of training samples, and $x$ is the sample for which we want to predict the binary class label. Note that the number of \gls{flop}s performed at inference is $4\cdot N\cdot d$, assuming that the \gls{flop}s  required to compute $k(x,x_i)$ is $2\cdot d$ (we consider only the dot-product between input vectors). If we now use in-memory kernel approximation, the predictive function becomes
\begin{equation*}
    f(x) = \text{sgn}(w^Tz(x))
\end{equation*}
where $w = \sum_{i=1}^{N}{\alpha_i z(x_i)}$. Since the matrix-vector multiplication required to compute $z(x)$ for in-memory kernel approximation is performed in analog, the total number of \gls{flop}s executed on digital hardware to perform the inference step for the single sample is equal to $2\cdot D$, where $D \ll N$.

\begin{figure}[h]
\includegraphics[width=\textwidth]{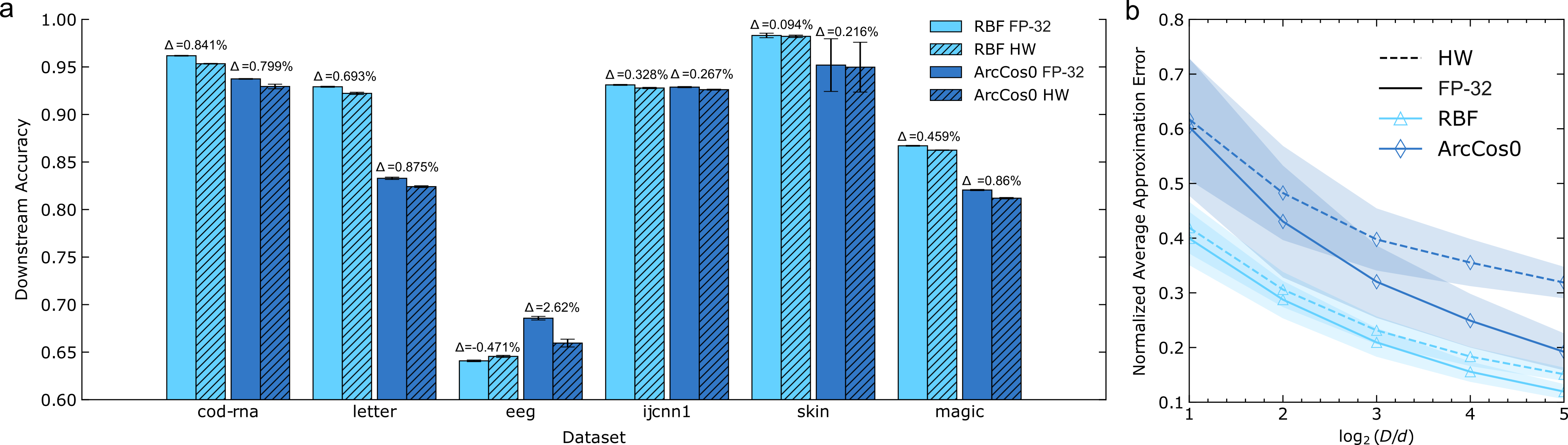}
\caption{\textbf{Experimental results for the hardware-accelerated approximations of the RBF and arc-cosine kernel}.
\textbf{a} Downstream classification accuracy experimental results. A remarkable retention of performance is obtained when the kernel approximations are deployed on the IBM HERMES Project chip, with an average empirical accuracy loss of $0.481\%$ for the \gls{rbf} kernel and $0.939\%$ for the zeroth-order arc-cosine kernel.
Only one kernel (arc-cosine) on a single benchmark (EEG) shows an accuracy delta greater than $1\%$, while all the other results show a smaller gap, with almost half of them losing less than $0.5\%$ accuracy in \gls{aimc} hardware compared to the equivalent floating-point baseline.
We include the delta $\Delta$ between the FP-32 precision and \gls{aimc} hardware implementation for each combination of kernel and dataset, $\Delta=\text{acc}(\text{fp})-\text{acc}(\text{hw})$.
Each bar in the plot is averaged over different random seeds ($10$) and approximation techniques (\gls{rff}, \gls{orf}, and \gls{sorf}), obtained for a fixed $\log(D/d)=5$. The error bars indicate the standard deviation calculated across the different random seeds, and averaged across the approximation techniques.
\textbf{b} Comparison between the normalized approximation error in the FP-32 and the \gls{aimc} hardware implementations. 
The plotted error is obtained by normalizing the approximation error for each task by the maximum error obtained across different approximations and benchmarks on that same task, and then averaging it across different tasks. 
As expected, we observe an increase in the normalized approximation error measured for the models deployed on the \gls{aimc} hardware, which is particularly noticeable for higher log-ratios on the zeroth-order arc-cosine kernel, but also present to a lower extent in the experiments on the \gls{rbf} kernel.
All the standard deviations in \textbf{a} and \textbf{b} are reported over $10$ different random seeds.
The results on the downstream classification accuracy and the approximation error are reported in detail in \ref{sup:extendedresults}.
}
\label{fig:kernelres}
\end{figure}

In Fig.~\ref{fig:kernelres}, we compare the downstream accuracy of a kernelized ridge classifier, as well as the approximation error, between in-memory kernel approximation (referred to as HW) and the implementation on conventional FP-32 software. For details on how we deploy the classifier on the \gls{aimc} chip, see Methods. The experiments were repeated $10$ times for the Gaussian and ArcCos kernel. For every kernel, we averaged the result across different sampling strategies, namely \gls{rff}, \gls{orf}, and \gls{sorf} (see \ref{sup:extendedresults} for extended results). Fig. \ref{fig:kernelres}a shows that we are within 1\% of the downstream accuracy on all tasks, for all kernels, except for the ArcCos kernel on the EEG classification task, where the difference is 2.62\%. 

The reported results were obtained for a ratio $\text{log}_2(D/d)=5$, i.e., the dimension of the concatenated vectors after mapping is $32\times$ the dimension of the input vector. This means that for the Gaussian kernel $m=16 \cdot d$ and for the ArcCos kernel $m=32 \cdot d$. To verify that increasing the dimension has the desired effect of reducing the approximation error, we conducted an experiment where we gradually increased the hidden dimension $D$ and recorded the approximation error. Fig.~\ref{fig:kernelres}b shows that the \gls{aimc} hardware is generally capable of approximating the true kernel matrix. We observe that in the regime of very high $D$, where the benefit of sampling more vectors $\omega$ starts to diminish, the gap between the \gls{aimc} hardware and FP-32 widens, especially for the ArcCos kernel. This can be attributed to the various sources of nonidealities present in the \gls{aimc} hardware for which we have not used any mitigation or compensation techniques. Note that the purpose of these experiments is to rather evaluate the intrinsic robustness of various kernel approximation methods in the presence of such nonidealities (see Methods).
From these experiments we conclude that if the highest approximation quality is required for a specific application (see the regime in Fig.~\ref{fig:kernelres} where $D \gg d$), one should resort to the digital solution. However, as we show in Fig. \ref{fig:kernelres}a, this is not necessary for a variety of classification tasks we study in this section. For a general overview of the computational cost of various methods that allow to project the input samples to high-dimensional hyperspaces for kernelized ridge classification, see Supplementary Table~\ref{tab:complexity}.

\subsection{Kernel-based attention in linear Transformers.}
As a second application, we investigate in-memory kernel approximation for approximating the Softmax computation\cite{choromanski2020rethinking} in the attention mechanism of Transformers.
The Transformer\cite{attentionisallyouneed} is a special neural network architecture based on the attention mechanism, and has established itself as the de facto standard in many domains such as natural language processing~\cite{chen2018,luo} and generative modelling~\cite{parmar}. Each attention layer comprises multiple so called attention heads, where $n_\text{head}$ is the number of heads and $d_\text{head}$ is the vector size in each head. Typically, the embedding dimension $d_\text{model}$ equals the head dimension multiplied by the number of heads. For each head, the input sequence $X \in \mathbb{R}^{L \times d_\text{model}}$, with $L$ being the sequence length, gets projected to a key $K$, a query $Q$, and a value $V$ matrix, each of size $L \times d_\text{head}$. For a given query vector $q$, we compute $L$ scores (one for each key vector $k$), representing the amount of attention we pay to each token. This is done for every query, yielding a matrix of shape $L \times L$, which is called the attention matrix. For every head $i$, the attention mechanism can be written as
\begin{equation} \label{eq:attention}
    \text{Attn}(Q,K,V)=\text{Softmax}\left(\frac{Q \cdot K^T}{\sqrt{d_\text{head}}}\right) \cdot V
\end{equation}
Unfortunately, as Eq.~\eqref{eq:attention} shows, the computational cost of the attention mechanism grows quadratically in $L$, which is problematic considering that many tasks such as summarizing or translating text need to process very long sequences. As a result, servicing models capable of processing long sequence lengths becomes prohibitively expensive, or even impossible when compute resources are scarce.

Naturally, this problem has motivated researchers to come up with cheaper alternatives, for which a comprehensive taxonomy is outlined in a recent review paper\cite{linearreview}. One of the alternatives revolves around the idea of viewing the Softmax in the attention mechanism as a kernel that can be approximated in an unbiased fashion using kernel methods\cite{transfrnns,choromanski2020rethinking,peng2021random,zhen2022cosformer}. As a result, the order in which the attention mechanism computes partial results can be changed, leading to an overall linear time-complexity that is for example exploited in the Performer~\cite{choromanski2020rethinking} architecture that we study in this manuscript.

\begin{figure}[h]
\includegraphics[width=\textwidth]{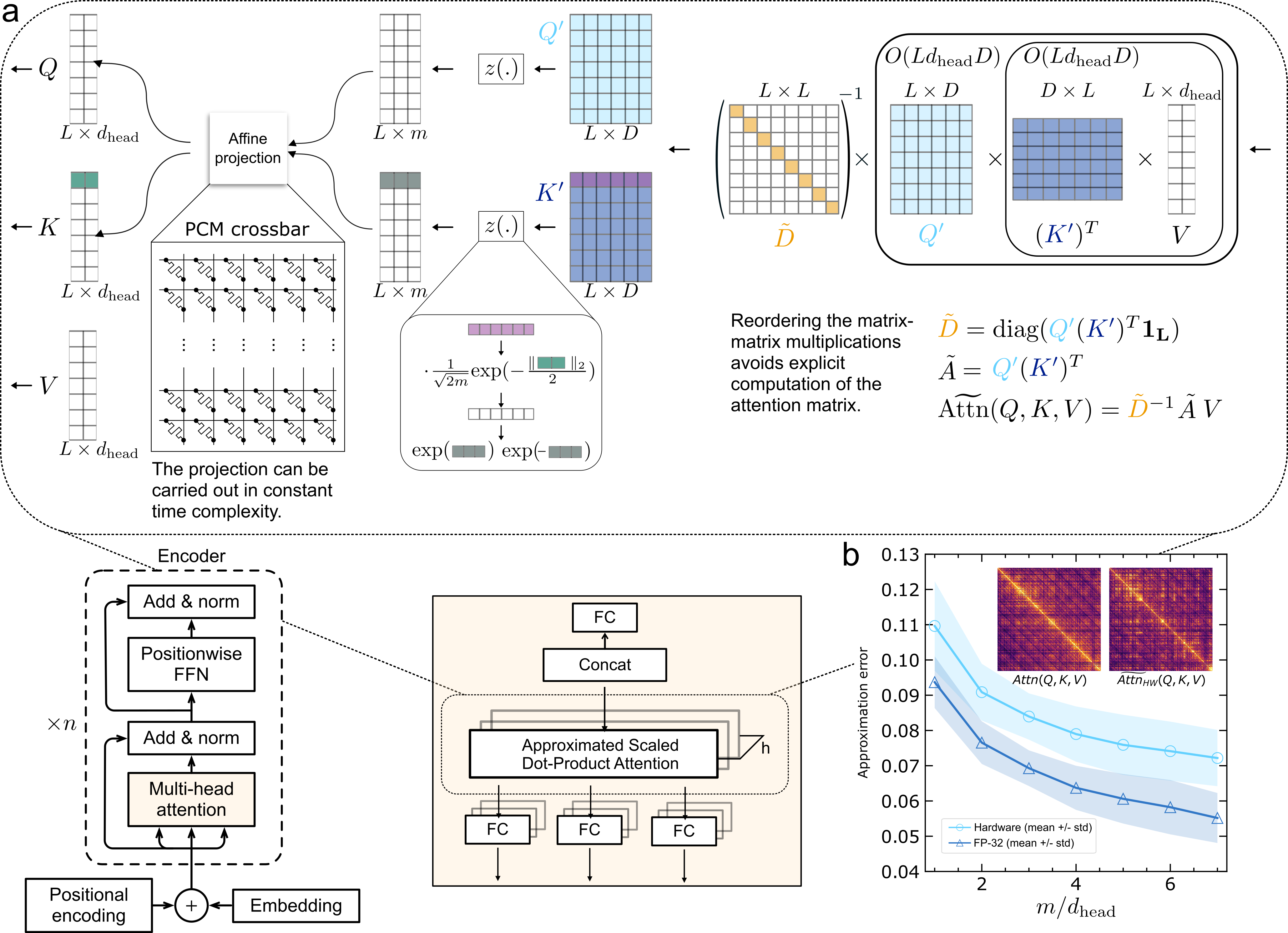}
\caption{\textbf{Schematic overview of in-memory kernel-approximation for \gls{mha}.} \textbf{a} After projecting the input to the query, key and value matrices for each head, they are fed into the "Approximated Scaled Dot-Product Attention" units, where $h$ is the number of heads. Inside each unit, the query and key vectors are projected into the $m-$dimensional space using \gls{aimc}, followed by the Softmax kernel specific post-processing $z$. Finally, the attention output can be calculated without explicitly calculating the attention matrix, due to the ability to re-order the matrix-matrix multiplications. This is possible because calculating the Softmax function over the attention scores is not necessary anymore.
\textbf{b} Experimental results on the IBM HERMES Project chip show that approximating the Softmax kernel results in slightly higher approximation errors compared to FP-32. Moreover, the approximation error (measured as the distance to the exact attention matrix) becomes smaller as we increase the hidden dimension $m$.}

\label{fig:performer}
\end{figure}

The process of approximating the Softmax kernel is depicted in Fig. \ref{fig:performer}a. For each head in the multi-head attention computation, the $d_\text{head}$-dimensional key and query vectors are first projected into the $m$-dimensional space using the feature vectors specific to the Softmax kernel. After mapping, the vectors are post-processed by first concatenating the positive and negative exponential of each $m$-dimensional vector, yielding $D$-dimensional vectors which are then passed through a Gaussian function. After having computed the projected versions of $Q$ and $K$, denoted as $Q'$ and $K'$, one can compute the attention output as $\tilde{D}^{-1}(Q'(K'^TV))$, where the brackets indicate the order of computation and $\tilde{D}$ is a diagonal matrix that contains the normalization values for the Softmax. By re-ordering the computation in this way, we avoid computing the attention matrix and reduce the computational complexity to $\mathcal{O}(LdD)$. In terms of \glspl{flop}, we can see that if $D=2 \cdot m$, the mapping accounts for roughly one third of the total \glspl{flop} of the multi-head attention computation, making it worthy of acceleration using \gls{aimc}.

Since directly evaluating the downstream accuracy of a Transformer using the approximate in-memory attention could obfuscate the effect of the analog noise on the approximation error, we first study the attention approximation error in an isolated manner. For this, we extracted query, key and value matrices from a random layer in an encoder-only Transformer (see Methods for details) and compared the on-chip approximation against the approximation done in FP-32 (see Fig.~\ref{fig:performer}b). We observe similar trends as in the previous kernel experiments where the analog noise increases the approximation error slightly, but the induced gap in approximation error stays constant as the hidden dimension $m$ is increased.

After having established that the analog noise increases the approximation error only slightly, we can move on to a benchmark commonly used to evaluate linear-complexity attention-approximation methods in Transformers, the \gls{lra}. The \gls{lra} is a collection of five text- and image-based tasks, where the inputs have long sequence lengths (up to 8,000 tokens) with long-range dependencies. Long-range dependencies ensure that methods relying solely on the locality of the inputs would fail to produce competitive results (see Methods).

Table \ref{tab:aaa} shows that, when performing the mapping on-chip, we incur \textit{no loss} in accuracy compared to the FP-32 baseline (denoted $\text{Performer}^\text{Vanilla training}$). Furthermore, we achieve an average score on-par with the state-of-the-art results reported\cite{chen2021skyformer}, while offloading between half and one third of the \glspl{flop} involved in the attention computation to \gls{aimc}. Here, $m = 4 \cdot d_\text{head}$ for all tasks except IMDb, for which $m = 8 \cdot d_\text{head}$. Interestingly, we did not have to do any hardware-aware finetuning of the networks\cite{Joshi2020,buechel-iclr,rasch-natcomm,murray1994}, which can be attributed to the fact that the random mapping matrices are especially resilient to noise as they are periodically re-drawn during training (see \ref{sup:training}). Additionally, we observe that the mapping matrices can be shared across layers, therefore incurring only constant memory overhead, instead of linear in the number of Transformer layers.

Finally, we perform additional experiments where all \glspl{mvm} involving static weights are done on-chip in addition to the attention mapping. In this case, a larger degradation in performance is observed for every \gls{lra} task ($2.52\%$ on average). For these networks, we found that strong hardware-aware training techniques had to be employed to obtain such accuracy on hardware (see Methods). We observe that some layers, especially the last layer, have very few parameters while being crucial for good accuracy. Concretely, performing the \gls{mvm} of the last layer in FP-32 for the Retrieval and Pathfinder task boosts performance by 1.55\% and 3.2\%, respectively. Additionally, we find that these networks are generally under-parameterized and expect that increasing the number of parameters would yield higher robustness. On IMDb and Listops, however, we perform very well, even achieving a slightly higher performance on IMDb. We attribute this to statistical noise because only one run was performed in hardware for these experiments.

\clearpage

\section{Discussion}
Approximation techniques based on random features offer a simple and efficient framework to approximate kernel functions with strong mathematical guarantees. The Monte Carlo approach adopted by these data-independent techniques allows to potentially average out sources of noise in the sampled feature vectors. In our experiments, we observe that the approximation error using \gls{aimc} follows a similar trend to the error when the approximation is performed in FP-32. We, however, observe that when the dimension of the feature space grows larger, the approximation using \gls{aimc} saturates, while the approximation in FP-32 continues to benefit from the higher number of features. Nevertheless, we find that the downstream performance of various benchmarks are resilient to this increased gap in the approximation error as we demonstrate on a wide number of tasks for a range of kernels. We demonstrate iso-accuracy on many kernelized ridge classification benchmarks, as well as iso-accuracy on the \gls{lra} benchmark when deploying the kernel approximation in the \gls{aimc} hardware.
We further demonstrate that no hardware-aware finetuning for the attention of Transformers is necessary when deploying the Softmax kernel approximation on-chip, which we attribute to the inherent robustness of the models to different realizations of the mapping matrices. This robustness is a direct consequence of periodic re-sampling of the feature vectors during training, which, as we show, is also crucial for generalization to the test set of each benchmark. Finally, we also demonstrate that we can achieve competitive performance on the \gls{lra} benchmark when all the stationary weights of the neural network are deployed on \gls{aimc}. However, in order to achieve this, hardware-aware finetuning of the models was necessary.

While the Softmax kernel yields, compared to the vanilla attention, competitive results, it comes with some non-negligible overhead in the form of operations such as element-wise multiplication, calculating the exponential and calculating the L2-norm, all of which need to be executed in a digital processing unit. Furthermore, the dimension $m$ to which one projects using \gls{aimc} hardware is only half of the target dimension $D$, meaning that we perform "only" $2 L d m$ \glspl{flop} on-chip while the remaining $2 L d D$ \glspl{flop} need to be carried out in FP-32. It has been shown multiple times~\cite{dissection,zhen2022cosformer} that the non-negativity of the attention matrix, which is the result of applying the Softmax function to $QK^T / \sqrt{d_\text{head}}$, is one of main reasons for the good performance of the vanilla attention mechanism. We therefore tested a simplified attention mechanism where $\tilde{\text{Attn}}(Q,K,V)=\tilde{D}^{-1} Q' (K')^T V$ with $Q'=\text{ReLU}(Q \Omega)$, $K'=\text{ReLU}(K \Omega)$ and $\tilde{D}=\text{diag}(Q'(K')^T \bf{1_L})$. We find that on the Cifar10 task of the \gls{lra} benchmark, possibly due to the lack of exponentials, training is more stable and we converge to an FP-32 test accuracy of 48.83\% while achieving 45.95\% when all the stationary weights are deployed on-chip (improvement of 3.25\% with respect to the results obtained using the Softmax kernel). We note that in this case, the on-chip mapping matrix $\Omega \in \mathbb{R}^{d \times D}$ directly maps into the $D$-dimensional space. Using this variant of the attention mechanism allows us to offload half of the total \glspl{flop} involved in the attention mechanism to \gls{aimc}, while gaining better performance compared to the Softmax version.

Although modern GPUs have a high peak throughput, our proposed in-memory kernel approximation scheme presents a way to perform kernel approximation with low latency, and especially low energy consumption.
With a peak throughput of 312 \gls{tops} at FP-16 precision, GPUs are hard to compete with. However, as we also discuss in \ref{sup:comparison-to-digital}, these peak throughput numbers are often only reached at large problem sizes. The same holds for the IBM HERMES Project Chip. The peak throughput of 63.1 \gls{tops} is only reached when all 64 cores are doing \glspl{mvm} in parallel. In our experiments, we often under-utilize our hardware. We however note that increasing the throughput on our chip is easy: one can simply replicate the mapping matrix across different cores and perform the mapping on those cores in parallel. If one would like to increase the throughput beyond 63.1 \gls{tops}, this is also possible by utilizing multiple chips in parallel. This of course increases the footprint significantly. We, however, note that the footprint of the IBM HERMES Project Chip is $144 mm^2$ while it is $826 mm^2$ for the NVIDIA A100. Additionally, in the context of kernelized attention, the mapping weights can be shared across all layers in the Transformer, meaning that the footprint attributed to those weights is constant. Note that this also holds on digital hardware, however, these weights still need to be fetched from memory owing to the von Neumann architecture.
Even if one assumes that the GPU achieves peak throughput, one key limitation of GPUs remains. With 400W peak power consumption and measured 70W static power consumption, the NVIDIA A100 is between 6.2 to 12.4 times less energy efficient compared to the IBM HERMES Project chip, which is running at a peak power consumption of 6.5W. Therefore, \gls{aimc} presents a viable option for running kernel approximation with low latency, and especially with low power consumption.
It should be noted that the hardware used in this paper is still prototypical and there are many ways to improve the energy efficiency and throughput of AIMC-based heterogeneous accelerators.
In the IBM HERMES Project Chip, the \gls{nvm} devices are integrated in the back-end-of-line of the underlying CMOS process. The devices are integrated several metal layers further from the underlying CMOS transistors. Integrating the devices closer to the transistors would significantly reduce the footprint of the crossbar array. As a result, more devices fit in one crossbar and the throughput can be increased. It is also possible to embrace \gls{nvm} devices such as Flash memory and ferroelectric field-effect transistors that do not require back-end-of-the-line integration.

Another way of increasing the number of devices per crossbar is by increasing the number of rows and columns. Adding more rows and columns yields a quadratic increase in the number of devices with only linear increase in the number of digital periphery circuits such as \glspl{dac} and \glspl{adc}. However, one also has to consider that if a crossbar is too big, the crossbar utilization decreases when the layers in a neural network do not occupy the whole crossbar. Another promising avenue to significantly increase the capacity of the computational tile without increasing the footprint is via 3D integration of the \gls{nvm} devices~\cite{Li2017,Wu2017}. The footprint and energy consumption of the tiles could be also reduced by implementing the digital periphery such as \glspl{dac} and \glspl{adc} in a more advanced node compared to the 14nm node that was used to fabricate the IBM HERMES Project Chip.

Besides increasing the number of devices per crossbar in order to increase the throughput, one can also further improve the latency of a single \gls{mvm}. This can be achieved by reducing the precision used to encode the inputs or by using a different input encoding scheme, such as the bit-serial scheme, and reducing the pulse duration of every bit slice. Furthermore, reducing \gls{adc} conversion time can further reduce latency.

Finally, note that high write endurance of the \gls{nvm} devices is not critical for in-memory kernel approximation, because the projection matrices are only programmed once and do not change during inference execution. The endurance in excess of $10^9$ erase-program cycles shown by \gls{pcm} devices~\cite{chen-pcm} is sufficient for this application, and could even be relaxed in favor of improving other parameters, such as reducing temporal conductance fluctuations.

To summarize, we propose the concept of in-memory kernel approximation. The essential idea is to leverage in-memory mapping operations to substantially improve the computational efficiency of linear projections associated with kernel approximation. Hence, in-memory kernel approximation presents an effective way to implement nonlinear functions using \gls{aimc}. We experimentally validated this scheme using the IBM HERMES Project Chip, a state-of-the-art in-memory computing chip based on \gls{pcm}. The efficacy of the approach was demonstrated for a traditional ridge classification model as well as a more recent linear-complexity transformer neural network model. The critical insight provided by our work, namely the on-chip deployment of kernel approximation techniques, facilitates further deployment of inference workloads for a wide range of machine learning tasks using \gls{aimc}.
Furthermore, our work provides an important demonstration in the area of linear-complexity attention in Transformers, as it represents the first \gls{aimc} deployment of the stationary weights of a Transformer for long-sequence modeling, extremely useful in a variety of research areas in other disciplines, such as Biology, Chemistry, and Physics.

\clearpage
\section*{Methods}
\subsection*{Datasets} \label{methods:datasets}
\subsubsection{Datasets used for the Kernel Approximation Experiments}
We consider six well-known \gls{ml} datasets to evaluate our implementations of standard kernel approximation, namely {IJCNN01}\cite{939502}, {letter}\cite{misc_letter_recognition_59}, {magic04}\cite{misc_magic_gamma_telescope_159}, {EEG}\cite{misc_eeg_eye_state_264}, {cod-rna}\cite{uzilov2006detection}, and {skin}\cite{misc_skin_segmentation_229}.
These benchmarks predominantly consist of binary classification tasks (except {letter}, which is an instance of a multi-class classification problem) and consider a broad range of input signals, from continuous EEG measurement to simulated high energy gamma particle tracings.
When available, we exploit the default partition between training and testing samples, while for the remaining datasets we perform a $50/50$ random partition of the data .
The absence of model selection in our study renders the validation splits useless, and we therefore merge them with the test split.
All datasets are normalized to zero mean and unit variance, as this reduces the error introduced by their quantization when fed into the hardware.

Additionally, we craft a synthetic dataset to evaluate the approximation error of the Softmax kernel used in FAVOR+ to approximate the attention mechanism.
This dataset is built by sampling key, query and value matrices for 50 different input sequences of variable length from the second encoder layer of a Performer model that was hardware-aware trained on one of the \gls{lra} tasks (IMDb).
More detailed information about the datasets, including their size, splitting method, and number of original features, are reported in Supplementary Table \ref{tab:dataker}.

We provide detailed replication experiments in \ref{sup:replication} in order to validate the correctness of our implementation of FAVOR+ and the kernel approximation used for ridge classification.

\subsubsection{Datasets used for the Perfomer Experiments}
We consider a widely used long-sequence benchmark to evaluate linear-complexity Transformer models, the \gls{lra}\cite{tay2021long}, because it challenges the model's ability to process large input sequences.
\gls{lra} is a collection of tasks comprising sequences ranging from one to 8 thousand tokens, encompassing a wide range of data types and modalities such as text, natural and synthetic images, and mathematical expressions requiring similarity, structural, and visual-spatial reasoning.
The benchmark includes five different datasets: ListOps, IMDb, ACL Anthology Network (AAN), CIFAR-10, and Pathfinder. More detailed information about the datasets are reported in Supplementary Table \ref{tab:dataperf}.
Altogether, they challenge a broad range of capabilities of the evaluated architectures, from modeling hierarchically structured data in long-context scenarios to long-sequence compression for similarity-based matching. This benchmark is also particularly fit for the investigation, as the models trained on it are relatively small (most of them perform well also with less than a million parameters) and every new linear-complexity Transformer variant is usually benchmarked on it, making it easier to compare with previous models in the literature. In this case, no re-scaling or normalization is applied to the input data, as the input samples are not fed directly into the hardware and hence they would not reduce the error introduced by quantization.
We use the standard training, validation, and test splits of the benchmark.

\clearpage
\subsection*{Model Training} \label{methods:model_training}
The investigated approximation algorithms do not require any form of training to approximate the true kernel function.
However, the algorithms used to tackle the downstream task, which, in our experiment for the standard kernels, corresponds to binary and multi-class classification, need to be trained on the task-specific extracted feature vectors.
In our experiments we leverage a proprietary PyTorch\cite{pytorch} implementation of the ridge classifier, a simple linear classification model based on ridge regression\cite{hoerl1970ridge} with closed form solution $w=(X^TX+\lambda I)^{-1}X^Ty$, where $X\in \mathbb{R}^{n\times m}$ is the training data matrix, $y\in \mathbb{R}^n$ is the training labels vector, and $\lambda\in\mathbb{R}$ is the regularization term.
This classifier is fit directly on top of the noise-free feature vectors $\mathbf{z}(x)$ extracted from each input sample $x$.
Since our goal is to verify to what extent these approximation techniques are robust to the noise introduced on hardware, we do not apply any form of hardware-in-the-loop or hardware-aware training, and use the FP-32 precision feature mappings to train the classification models which are later used to perform inference on feature vectors computed on-chip.
The regularization term is fixed to $\lambda=0.5$ across all the datasets; we do not engage in extensive hyper-parameter tuning activities, because with this value we are able to match previous results on the studied kernels\cite{liu2021random}
The training of these models is performed on standard consumer hardware, as it does not require excessive amounts of resources and time.

On the other hand, the Performer models investigated in our experiments on linear-complexity Transformers require a greater effort in terms of training time, resources and infrastructure compared to the classification models discussed so far.
Our re-implementation of these models leverages FairSeq\cite{ott2019fairseq}, an open-source sequence modeling toolkit from Meta AI that allows researchers and developers to train custom models for translation, summarization, language modeling, and other text generation tasks. 
FairSeq supports the integration of another toolkit from Meta AI, xFormers\cite{xFormers2022}, that contains a wide range of customizable and domain-agnostic components such as various efficient implementations of the attention mechanism, including FAVOR+ kernelized attention.
We adopt the model definition from previous works on linear-complexity Transformers\cite{chen2021skyformer}, to allow for a fair comparison in terms of model size and training procedure.
The trained models are relatively small, at most two encoder layers and 200 thousand trainable parameters, which is also convenient for hardware deployment.
All the models were trained from scratch on each dataset’s training partition for 50 thousand steps except for IMDb, where the training was limited to 20 thousand because of the fast convergence of the model, and Pathfinder, where the training was extended to 100 thousand steps since the learning rate had to be lowered to guarantee a more stable convergence to the target.
More details on the task-specific hyper-parameters used in our experiments can be found in Supplementary Table \ref{tab:hyperpar}.

During training, we empirically observe the importance of one of the hyper-parameters in particular: the re-sampling frequency for the feature vectors in the kernelized attention mechanism.
When they are not re-sampled during training, the model starts overfitting to the fixed mapping matrix and cannot extrapolate the learned solution when a different mapping matrix is plugged in.
Interestingly, feature re-sampling makes the model intrinsically robust to variations of the mapping matrix, allowing to plug in arbitrary matrices at inference time with no drop in the model performance, as long as they are sampled according to the correct distribution.
We speculate that this behavior represents the reason behind the robustness of the models when only the kernelized attention mechanism is deployed on \gls{aimc} hardware.
For more information on the importance of re-sampling the mapping features refer to \ref{sup:training}.
While the robustness induced by feature re-sampling during training already allows to deploy the FAVOR+ mechanism on \gls{aimc} hardware without any loss in performance, the acceleration of the full Performer model would result in a significant accuracy drop. In order to increase the robustness of the model, we employed conventional hw-aware training methods\cite{rasch-natcomm,Joshi2020} using the IBM AIHWKIT\cite{aihwkit,using-aihwkit}. More precisely, we learned the input ranges for 8-bit input quantization, where we encouraged tight input ranges during training in order to boost the signal strength, which leads to a better \gls{snr} at the \glspl{adc}. To further improve robustness to noise resulting from poor \gls{snr}, we injected Gaussian noise with magnitude $\eta_\text{out}=0.1$ into the output of each \gls{mvm}. During training, we also clipped the weights to $\alpha=2.0$ standard deviations, ensuring that, during programming, no outliers are mapped to high conductance values. Additionally, we also injected Gaussian noise $\eta_\text{train}=0.12$ into the weights, where the magnitude is proportional to the dynamic range of the weights, i.e. $\zeta \sim \text{max}(|W|) \cdot \eta_\text{train} \cdot \mathcal{N}(0,1)$, where $\zeta$ is the noise vector added to the weights.

The training for all the models is carried out on a single NVIDIA A100 40GB \gls{gpu}.
The learning curves for the accuracy and loss for the trained models, as well as the evaluation of their best checkpoint (according to the validation accuracy) repeated and averaged across 10 different random seeds (that is, with 10 different FAVOR+ mapping matrices) on the test partition of each task are included in the \ref{sup:training}.

\subsection*{Evaluation Platform} \label{methods:platform}
In our experiments we use the IBM HERMES Project Chip\cite{mt-hermes}, which is an advanced \gls{aimc} chip fabricated using 14-nm CMOS technology. It features 64 cores, each core hosting a $256\times 256$ crossbar with backend-integrated \gls{pcm}. Each crossbar comprises 65,536 unit cells, each unit cell representing one synaptic weight. The unit cell is made up of four \gls{pcm} devices, representing positive and negative weights in a differential manner (2 devices per sign). The chip has a total weight capacity of $4,194,304$ programmable weights. Each core additionally hosts 256 \glspl{dac} that convert INT8 inputs to voltage pulses where the length of the pulse is proportional to the INT8 input value. In order to read out the currents that accumulate along the columns in parallel, 256 current-controlled oscillator-based \glspl{adc} are integrated into each core. For further affine corrections, each core also has digital circuitry to apply a scale and offset to the output vector in parallel.
The chip is mounted onto a hardware platform comprising a base board and an FPGA. The base board provides the infrastructure for hosting the chip, such as the socket, cooling, power supplies, as well as voltage and current reference sources. The FPGA is used for controlling the chip. This involves reading and writing control registers, sending and receiving data, as well as sending instructions. The communication between the FPGA and chip happens through a bi-directional serial interface. The hardware platform is connected to a local computer via Ethernet, and the chip can be controlled using a high-level Python-based library.

\subsection*{Deployment on the IBM HERMES Project Chip} \label{methods:deployment}
The deployment on the chip is performed using a dedicated software stack, that allows for end-to-end deployment of complex \gls{dl} inference workloads on this \gls{aimc} accelerator.
The software stack handles all the steps necessary for deploying artificial neural networks onto the chip, such as the symbolic tracing of the architecture, pipelining of the operations, the mapping of the weights onto the crossbar arrays, and the post-compilation accuracy tuning for the model.
In particular,
\begin{enumerate}
    \item During the symbolic tracing using {torch.fx}\cite{torch-fx}, proxy objects are forwarded through the model to compile the network in a graph structure. In the compiled network, each node represents an operation and contains information about its inputs, parameters, and outputs, while each edge represents dependencies between operations.
    \item After the network has been successfully compiled as a graph, a first round of course-grained operation pipelining is performed at the operation level of abstraction, that is, on the nodes of the graph; the operations are compiled from a graph structure into a “pipeline structure”, where operations are grouped into chains and branches. Chains are executed sequentially, while branches are executed in parallel (within the chain). The inputs of the branches are either the output values of the previous chain or cached values stored previously during the execution. During this step, operations that can be accelerated on hardware are identified. Executing the model in a pipelined fashion maximizes the number of parallel \glspl{mvm} performed on-chip.
    \item After compilation, 2,000 inputs from the training set are -- in emulated mode -- fed through the network and the quantized inputs into each crossbar are cached. Using the weights and cached inputs of every crossbar, we then calculate the maximum current per column flowing into each \gls{adc} as a function of the maximum conductance per column. This allows us to then determine the optimal column-wise maximum conductance we can use for mapping our weights to conductance levels without saturating the \glspl{adc}.
    \item Once this post-compilation tuning generated the weights to be mapped to the hardware, they are programmed to the physical crossbar arrays using \gls{gdp}\cite{msf,jetcas-buechel}.
\end{enumerate}
Furthermore, the software stack also handles the execution of the compiled model (both on real hardware and in emulated mode, where the different noise components are simulated with noise models that are specific to the IBM HERMES Project Chip) by interfacing itself with lower-level drivers, which handle the execution of the operations on the hardware.

\newpage
\section*{Data availability}
The data that support the plots within this paper and other findings of this study are available from the corresponding author upon reasonable request.

\section*{Code availability}
The code used to generate the results of this study is available at \\ \url{https://github.com/IBM/kernel-approximation-using-analog-in-memory-computing}.

\section*{Acknowledgments}
This work was supported by the IBM Research AI Hardware Center. This work has also received funding from the European Union's Horizon Europe research and innovation program under Grant Agreement No 101046878, and was supported by the Swiss State Secretariat for Education, Research and Innovation (SERI) under contract number 22.00029. We would also like to thank Thomas Hofmann for fruitful discussions.

\section*{Author contributions}
J.B. and G.C. set up the infrastructure for training and evaluating the various models. J.B., G.C., A.V. and C.L. set up the infrastructure for automatically deploying trained models on the IBM Hermes Project Chip.
J.B. and G.C. wrote the manuscript with input from all authors. A.R., M.L.G. and A.S. supervised the project.

\section*{Competing interests}
The authors declare no competing interests. 
\clearpage

\section*{Tables}
\newcolumntype{P}[1]{>{\centering\arraybackslash}p{#1}}
\newcolumntype{L}[1]{>{\raggedright\let\newline\\\arraybackslash\hspace{0pt}}m{#1}}
\newcolumntype{R}[1]{>{\raggedleft\let\newline\\\arraybackslash\hspace{0pt}}m{#1}}

\begin{table}[h]
\centering
\begin{tabular}{ p{5.5cm}P{1.8cm}P{1.8cm}P{1.8cm}P{1.8cm}P{1.8cm}P{1.8cm} } 
  \toprule
  \textbf{Model}  & \textbf{IMDb} & \textbf{Listops} & $\textbf{Retrieval}^*$  & $\textbf{Pathfinder}^*$  & \textbf{Cifar-10} & \textbf{AVG.} \\ 
\midrule
& \multicolumn{6}{c}{\textbf{Previously reported results\cite{chen2021skyformer}}} \\ 
 Transformer  &  61.95& 38.37 & 80.69 &65.26 &40.57& 57.37 \\
 Linformer & 58.93 & 37.45 & 78.19 & 60.93 & 37.96 & 54.69 \\
 Nystr\"{o}mformer & 64.83&  38.51&  80.52&  69.48 & 41.30&  58.93\\
 Skyformer & 64.70 & 38.69 & 82.06 & 70.73 & 40.77 & 59.39\\
 Performer &  64.19 & 38.02&  80.04 &66.30 &41.43 & 58.00 \\
 \midrule
& \multicolumn{6}{c}{\textbf{Our implementation}} \\ 
 $\text{Performer}^\text{Vanilla training}$ &  $66.09^{\pm0.99}$ & $37.61^{\pm0.08}$ &  $77.96^{\pm0.76}$  & $70.85^{\pm1.18}$ & $45.95^{\pm0.72}$ & $59.69^{\pm0.75}$\\
 $\text{Performer}^\text{Vanilla training}$ on-chip attn. only & $66.15^{\pm1.08}$ & $37.56^{\pm 0.16}$ & $77.51^{\pm 0.87}$ & $70.94^{\pm1.13}$ & $45.89 ^{\pm0.75} $ & $59.61^{\pm 0.80}$ \\
 $\text{Performer}^\text{HWA training}$ & $65.88^{\pm1.06}$ & $38.63^{\pm0.20}$ & $78.4^{\pm0.55}$ & $72.11^{\pm0.89}$ & $46.03^{\pm0.65}$ & $60.21^{\pm0.67}$\\
 $\text{Performer}^\text{HWA training}$ on-chip full model & $66.20$ & $38.30$ & $75.6$ & $65.65$ & $42.70$ & $57.69$\\
 \bottomrule
\end{tabular}
\caption{\textbf{Accuracy of the \gls{aimc} hardware-accelerated Performer model on the \gls{lra} benchmark.} 
\newline
$^*$ Accuracies for Retrieval and Pathfinder increase to $77.15\%$ and $68.85\%$ respectively when the last layer is deployed in FP-32 for the full on-chip experiments.
}
\label{tab:aaa}
\end{table}

\clearpage

\section*{Figures}

\clearpage

\clearpage

\section*{References}
\bibliography{curated_bib}

\clearpage
\renewcommand{\figurename}{Supplementary Figure}
\setcounter{figure}{0}
\renewcommand{\tablename}{Supplementary Table}
\setcounter{table}{0}
\renewcommand\thesection{Supplementary Note ~\arabic{section}}
\setcounter{section}{0}

\clearpage
\section*{Supplementary Tables}
\begin{table}[h]
\centering
\begin{tabular}
{P{1.1cm} P{4cm} P{1cm} P{1cm} P{1cm} P{4cm} P{3cm} P{2cm}} 
    \toprule
    \textbf{Kernel} & \textbf{Kernel definition} & $f_1(x)$ & $f_2(x)$ & $l$ & $h(x)$  & $p(\omega)$ & $b$  \\ 
\midrule
\gls{rbf}  &  $\text{exp}(-\|x-y\|_2^2 / 2)$  & sin & cos & 2 & 1 & $\sqrt{2} \cdot \mathcal{N}(0,1)$ & $2\pi \cdot \mathcal{N}(0,1)$ \\
ArcCos0 & $1-1/\pi \cdot \text{cos}^{-1} \frac{x^Ty}{\|x\|_2 \|y\|_2}$  & $\Theta (x)$ & -  & 1 & $\sqrt{2}$ & $\mathcal{N}(0,1)$ & 0 \\
Softmax & $\text{exp}(x^Ty)$ & $\text{exp}(x)$ & $\text{exp}(-x)$  & 2 & $\sqrt{2} \cdot \text{exp}(-\|x\|_2^2 / 2)$ &  $\mathcal{N}(0,1)$ & 0 \\
\bottomrule
\end{tabular}
\caption{\textbf{Kernel definitions.} The general definition of each kernel evaluated on vectors $x$ and $y$ according to equation \ref{eq:rff}. $\Theta$ refers to the heaviside function. To avoid outliers of $\Omega$, which would map to high conductance states on our hardware, while the other weights would map to very low conductances, we replaced every Gaussian distribution by a truncated Gaussian, typically truncated at three standard deviations.}
\label{tab:kernel_def}
\end{table}

\begin{table}[h]
\centering
\begin{tabular}
{P{4cm} P{4.0cm} P{2.5cm} P{4.0cm} P{2.5cm}}
    \toprule
    \textbf{Technique} & High-dimensional Mappings \leavevmode\color[HTML]{397BAB} $\phi(x)^T\phi(y)$ & Kernel Methods \leavevmode\color[HTML]{397BAB}$k(x,y)$ & Kernel Approximations \leavevmode\color[HTML]{397BAB}$z(x)^T z(y)$ & \gls{aimc} Deployment \leavevmode\color[HTML]{397BAB}$z(x)^T z(y)$ \\ 
\midrule
\textbf{Inference \gls{flop}s} & $4\cdot H \cdot d + 2\cdot H$  & $2\cdot d \cdot N$ & $4\cdot m \cdot d + 2\cdot D$ & $2\cdot D$ \\
\bottomrule
\end{tabular}
\caption{
\textbf{Working in high-dimensional spaces: the evolution of the computational cost of inference.} 
We include an outline of the computational cost at inference time over different techniques that allow to project the input samples to high-dimensional hyperspaces, to facilitate a better understanding of how the cost of these methods evolved over time and how our proposed model compares with previous techniques.
We order the different techniques based on computational cost, from top to bottom.
We reference different variables which are introduced in the main paper, namely: the number of training samples $N$, the dimension of the Hilbert space $H$, the original dimensionality of the input samples (number of features) $d$, the dimension of the space to which the input samples are projected in kernel approximation techniques $m$, and the total number of features used by kernel approximations techniques $D$, where $D>m$ when more than one post-processing function is used (as in the case of the \gls{rbf} kernel).
$k(\cdot, \cdot)$ represents the kernel function, $\phi$ represent the mapping function which is implicitly expressed by the kernel function, while $z$ represent the explicit feature map used in kernel approximation techniques.
}
\label{tab:complexity}
\end{table}

\begin{table}[h!]
    \hfill
    \begin{tabular}{ p{3cm} P{1.5cm} P{1.5cm} P{1.5cm} R{1.5cm} R{1.5cm}} 
      \toprule
      \textbf{Dataset} & $d$  & \textbf{Split} & \textbf{Classes} & \textbf{Train} & \textbf{Test} \\ 
      \midrule
      IJCNN\cite{939502} & 22  & fixed & 2 & 49,990 & 91,701\\ 
      EEG\cite{misc_eeg_eye_state_264} & 14& random  & 2 & 7,490 & 7,490 \\ 
      cod-RNA\cite{uzilov2006detection} & 8 & fixed & 2 & 59,535 & 157,413 \\ 
      magic04\cite{misc_magic_gamma_telescope_159} & 10 & random & 2 & 9,510 &  9,510 \\ 
      letter\cite{misc_letter_recognition_59} & 16 & fixed & 26  & 12,000 & 6,000\\ 
      skin\cite{misc_skin_segmentation_229} & 3 & random & 2 & 122,529  & 122,529 \\ 
      attention  & 64 & random & - & $4000$ & $4000$ \\ 
      \bottomrule
    \end{tabular} \hfill 
    \hfill
    \caption{Datasets used for the Kernel Approximation Experiments.}
    \label{tab:dataker}
  \vspace*{0.5cm}
    \hfill
    \begin{tabular}{ p{3cm} P{1.5cm}  P{1.5cm} P{1.5cm}  P{1.5cm} R{1.5cm} R{1.5cm} R{1.5cm}  } 
      \toprule
      \textbf{Task} & $d$  & \textbf{Split} & \textbf{Classes} & \textbf{Modality}  &  \textbf{Train} & \textbf{Valid} & \textbf{Test}  \\ 
      \midrule
      ListOps & 2,000  & fixed & 10 & text  & 96,000 & 2,000 & 2,000 \\
      IMDb & 4,000 &  fixed & 2 & text  & 25,000 & 25,000 & 25,000  \\
      AAN & 8,000 &  fixed & 2 & text  & 147,086 & 18,090 & 17,437  \\
      CIFAR-10  & 1,024  & fixed & 10  & image&  45,000 & 5,000 & 10,000   \\
      Pathfinder  & 1,024 & fixed & 2  & image  &  160,000 & 20,000 & 20,000  \\
      \bottomrule
    \end{tabular}\hfill
    \hfill
    \caption{Tasks of the \gls{lra} benchmark used for the Performer Experiments.}
    \label{tab:dataperf}
  \caption{\textbf{Additional datasets statistics.} For each entry, we report the number of training, validation and test samples (in the  kernel approximation experiments validation and test set are merged), the number of input features $d$, the training/validation split technique used (``random'' means that the data was randomly split in two equal partitions, training and test, while ``fixed'' means that the dataset already provided fixed partitions for training and test). Additionally, we include the input modality (image or text) for each task of the \gls{lra} benchmark.}
    \label{tab:datasets}
\end{table}

\begin{table}[h!]
\centering
\begin{tabular}{ p{3cm} P{2cm} P{2cm}P{2cm}P{2cm}P{2cm}} 
    \toprule
      \textbf{Parameter} & \textbf{AAN} & \textbf{Pathfinder}& \textbf{IMDb} & \textbf{ListOps} & \textbf{Cifar-10} \\ 
      \midrule
        batch size &  128  &  256 &  128  &  256 &  128 \\   
        lr scheduler & inverse sqrt  & linear decay & inverse sqrt & inverse sqrt &  inverse sqrt \\   
        warmup updates & 4000 & 1000 & 4000 & 1000 &  4000 \\ 
         activation fn & gelu & gelu & silu& gelu &  gelu \\ 
        $\text{classifier}_{in}$ & 128 & 64 & 64 & 64 &  64 \\ 
        $\text{classifier}_{out}$ &  512 &  256 & 128 & 128 &  128 \\ 
        dropout & 0.1  & 0.0  & 0.1 & 0.1 &  0.1 \\ 
        attention heads & 2 & 8  & 2 & 2 &  2 \\   
        embed dim & 64 & 128 & 64  & 64 &  64 \\ 
        ffn embed dim & 128 & 128 & 128 & 128 &  128 \\ 
        encoder layers & 2  & 1 & 2& 2 &  2 \\ 
        sampled features & 256  & 256 & 256 & 256 &  256 \\ 
        redraw steps & 1500 & 1500& 1500 & 1500 &  1500 \\  
        clip norm & 0.5 & 1 & 0.5& 0.5 &  0.5 \\ 
        learning rate & 1e-3 & 1e-4 & 6e-4& 1e-3 &  6e-4 \\ 
        max update & 50e+3 & 100e+3 & 50e+3& 50e+3 & 10e+3 \\ 
        optimizer & adam & adam & adam & adam &  adam \\ 
        adam betas & (0.9,0.98) & (0.9,0.999) & (0.9,0.98) & (0.9,0.98) & (0.9,0.98)\\
        adam eps & 1e-9 & 1e-6 & 1e-9 & 1e-6 & 1e-9 \\
        weight decay & 0.1 & 0.01 & 0.1 &  0.1 & 0.1 \\
        normalization & - & - & - & (0.5, 0.5) &  (0.48, 0.24) \\ 
    \bottomrule
    \end{tabular}
\caption{\textbf{Hyper-parameters used to train our Performer models on the different tasks of the \gls{lra} benchmark.} Two different activation functions are used for the different tasks, namely GELU\cite{hendrycks2016gaussian} and SiLU\cite{elfwing2018sigmoid}. The parameter \textit{normalization} refers to the pixel normalization that is performed on the input samples, which are normalized to zero mean and unit variance; the first normalization term refers to the mean (minuend), while the second normalization term refers to the variance (dividend). The parameter \textit{redraw steps} refers to the frequency how often the mapping matrices in the FAVOR+ kernelized attention are re-sampled; $1500$ means that the mapping matrix is re-sampled every $1500$ update steps in the optimization procedure. On the other hand, the parameter \textit{sampled features} refers to the number of columns in the mapping matrix used in FAVOR+.}
\label{tab:hyperpar}
\end{table}

\clearpage
\section{Extended hardware experiments}
\label{sup:extendedresults}
We report extended results on the hardware deployment experiments for standard kernel approximation techniques, broken down for individual benchmarks and approximation techniques.
We present the results for {cod-rna}, {EEG}, {IJCNN01}, {letter}, {magic04}, and {skin} in Supplementary Figs. \ref{fig:extendedhwcodrna}, \ref{fig:extendedhweeg}, \ref{fig:extendedhwijcnn}, \ref{fig:extendedhwletter}, \ref{fig:extendedhwmagic}, and \ref{fig:extendedhwskin}.
These results come from the same experiments from which we extract the charts shown in the main paper, with the only difference being the level of granularity used to display them.
Also, in these graphs we do not normalized the approximation error across different benchmarks.
The reported approximation error is computed as defined in the main body of the paper, namely
\begin{equation*}
    \text{Approx. Error} = \frac{\|\mathcal{G}-\hat{\mathcal{G}}\|_F}{\|\mathcal{G}\|_F}
\end{equation*}
The $x$-axis of the graphs represents the logarithmic-ratio between the number of sampled features $m$ and the number of original features of the inputs $d$.
Observe that $d$ is not fixed and varies across different datasets. The value of sampled features $m$ is task-dependent and can be derived as $m=2^{r}d$, where $r$ is the value of the log-ratio.
In all these figures, we leverage the following style scheme:
\begin{itemize}
    \item approximation technique: violet (random Fourier features, \gls{rff}), blue (orthogonal random features, \gls{orf}), green (structured orthogonal random features, \gls{sorf})
    \item used hardware: solid line (floating-point precision experiments), dashed line (\gls{aimc} experiments)
\end{itemize}

\begin{figure}[!ht]
    \centering
    \includegraphics[width=\linewidth]{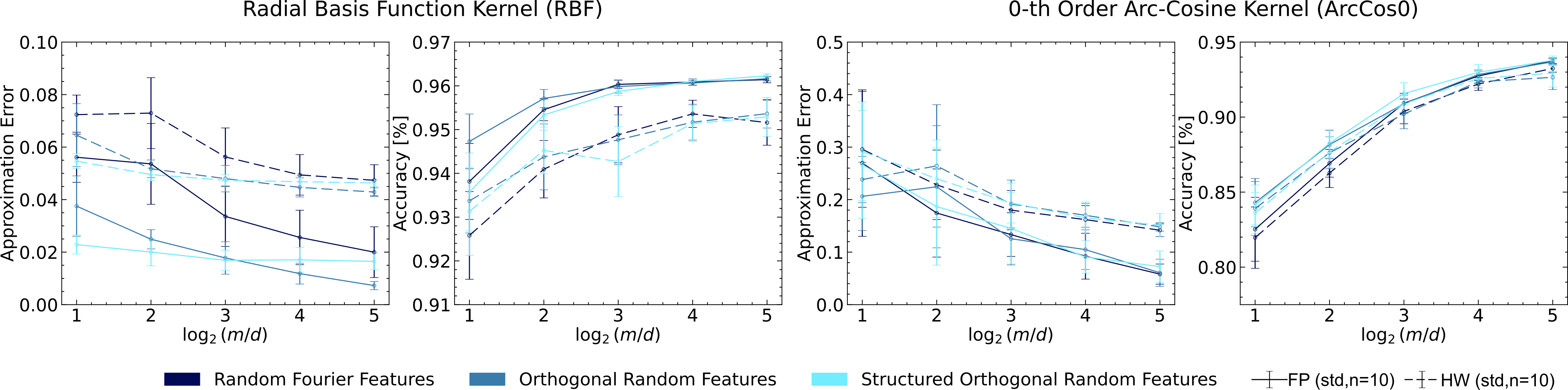}
    \caption{Hermes hardware experiments on the cod-rna dataset.}
    \label{fig:extendedhwcodrna}
\end{figure}

\begin{figure}[!ht]
    \centering
    \includegraphics[width=\linewidth]{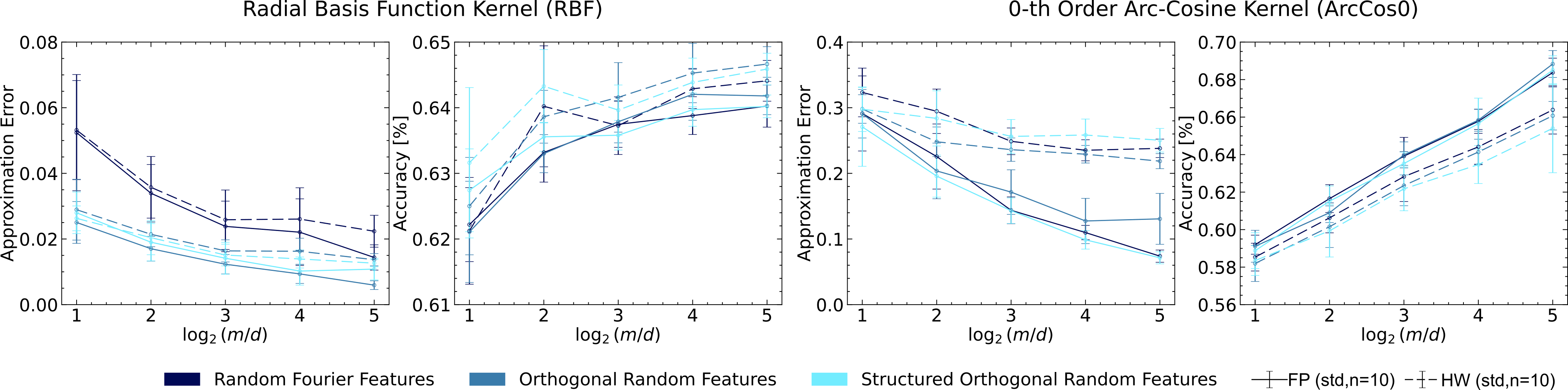}
    \caption{Hermes hardware experiments on the eeg dataset.}
    \label{fig:extendedhweeg}
\end{figure}

\begin{figure}[!ht]
    \centering
    \includegraphics[width=\linewidth]{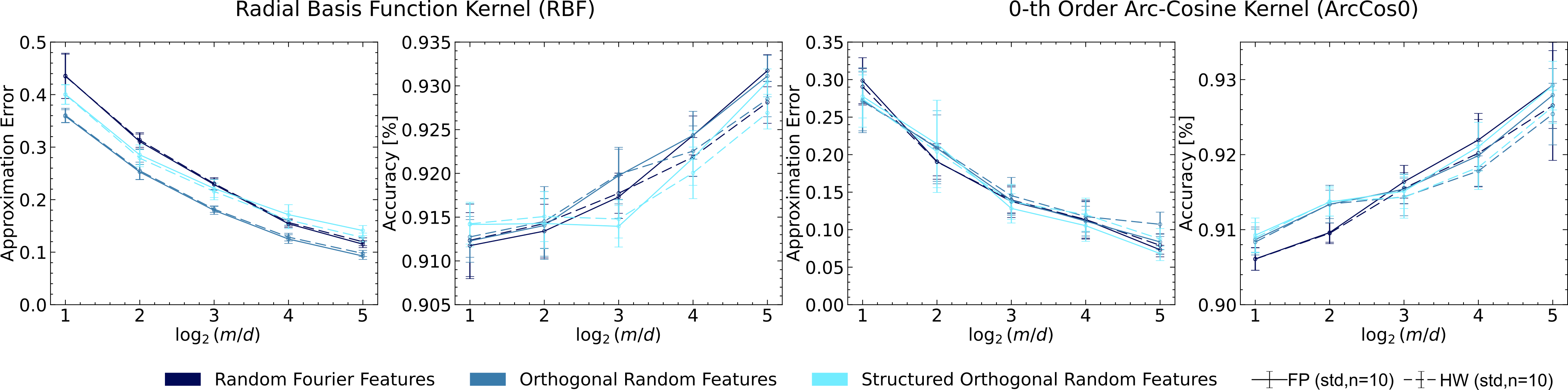}
    \caption{Hermes hardware experiments on the IJCNN1 dataset.}
    \label{fig:extendedhwijcnn}
\end{figure}

\begin{figure}[!ht]
    \centering
    \includegraphics[width=\linewidth]{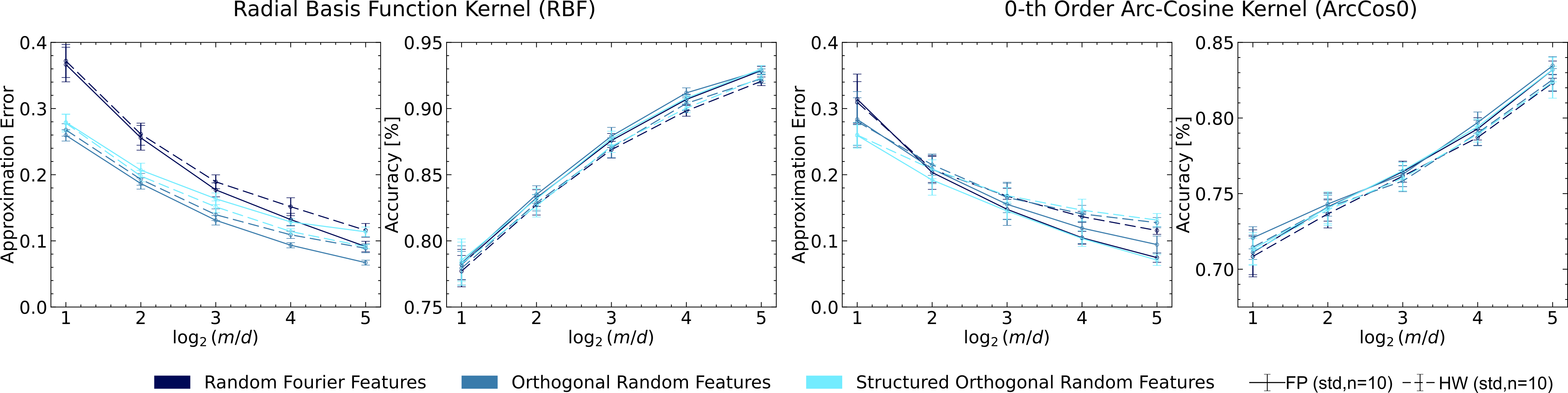}
    \caption{Hermes hardware experiments on the letter dataset.}
    \label{fig:extendedhwletter}
\end{figure}

\begin{figure}[!ht]
    \centering
    \includegraphics[width=\linewidth]{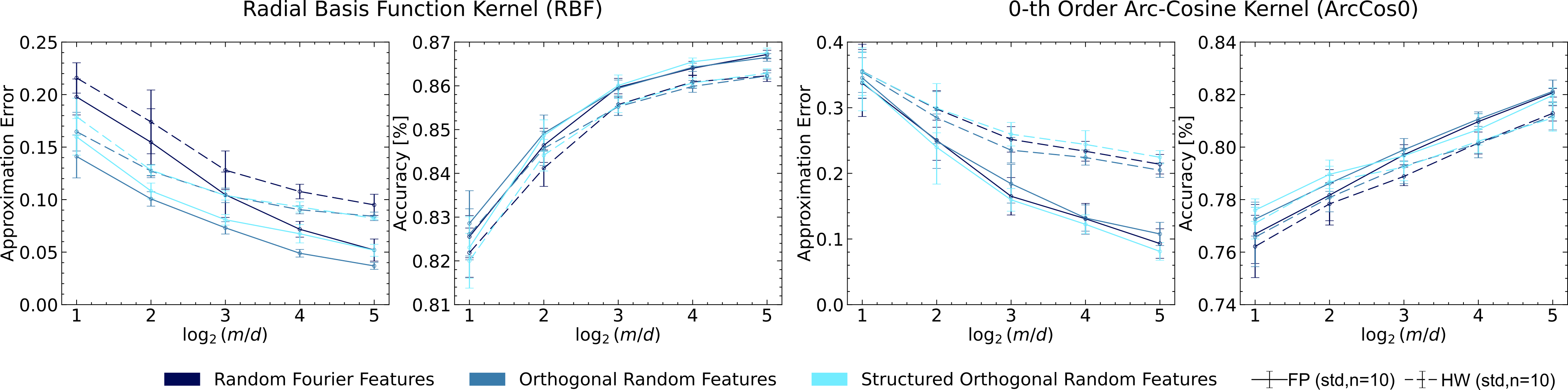}
    \caption{Hermes hardware experiments on the magic dataset.}
    \label{fig:extendedhwmagic}
\end{figure}

\begin{figure}[!ht]
    \centering
    \includegraphics[width=\linewidth]{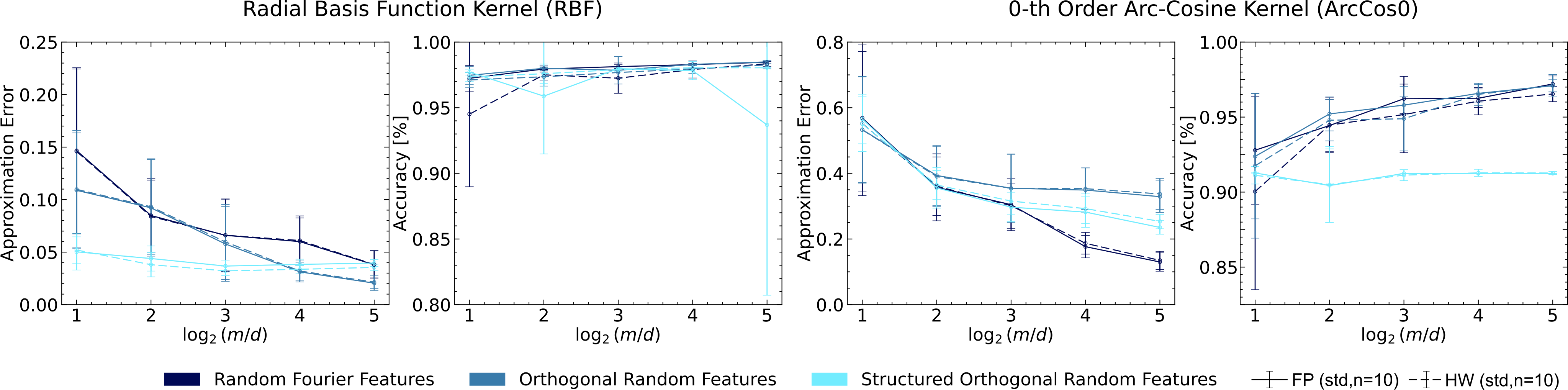}
    \caption{Hermes hardware experiments on the skin dataset.}
    \label{fig:extendedhwskin}
\end{figure}
\clearpage

\section{Performer model training details}
\label{sup:training}
Here, we report and discuss details on training the Performer models used in our experiments.
In Supplementary Table \ref{tab:hyperpar}, we show the specifics of the architectures trained, as well as the optimization parameters leveraged during training for each task of the \gls{lra} benchmark.
This setting of the Performer is actually inspired by a previous replication of the model that was done in \textit{Skyformer: Remodel Self-Attention with Gaussian Kernel and Nyström Method}\cite{chen2021skyformer}.
In this work, the Performer and many other linear-complexity models were re-implemented with comparable architectures (similar number of trainable parameters) to allow for a fair comparison between models.

Additionally, we provide the learning curves for our standard trained models.
In particular, we show both the measured accuracy and loss (training and validation) for the standard models in Supplementary Fig. \ref{fig:fptrainingacc} and \ref{fig:fptrainingloss}, respectively.

\begin{figure}[h!]
\centering
\begin{minipage}{0.3\textwidth}    
    \includegraphics[width=\textwidth]{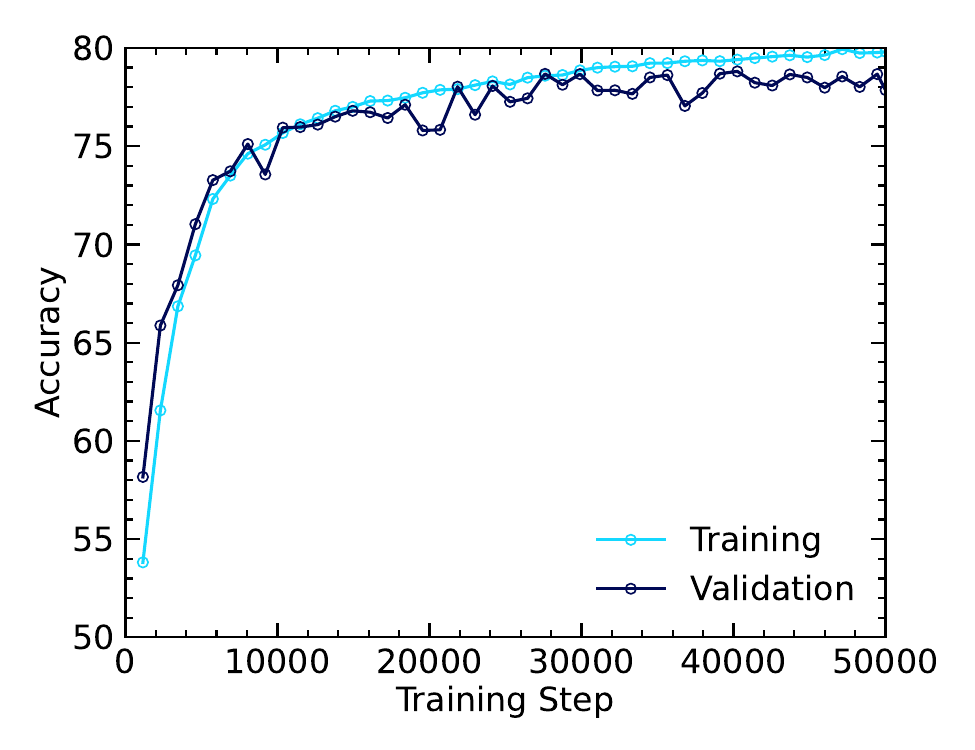}
    \caption{AAN task.}
\end{minipage}
\hfill
\begin{minipage}{0.3\textwidth}
    \includegraphics[width=\textwidth]{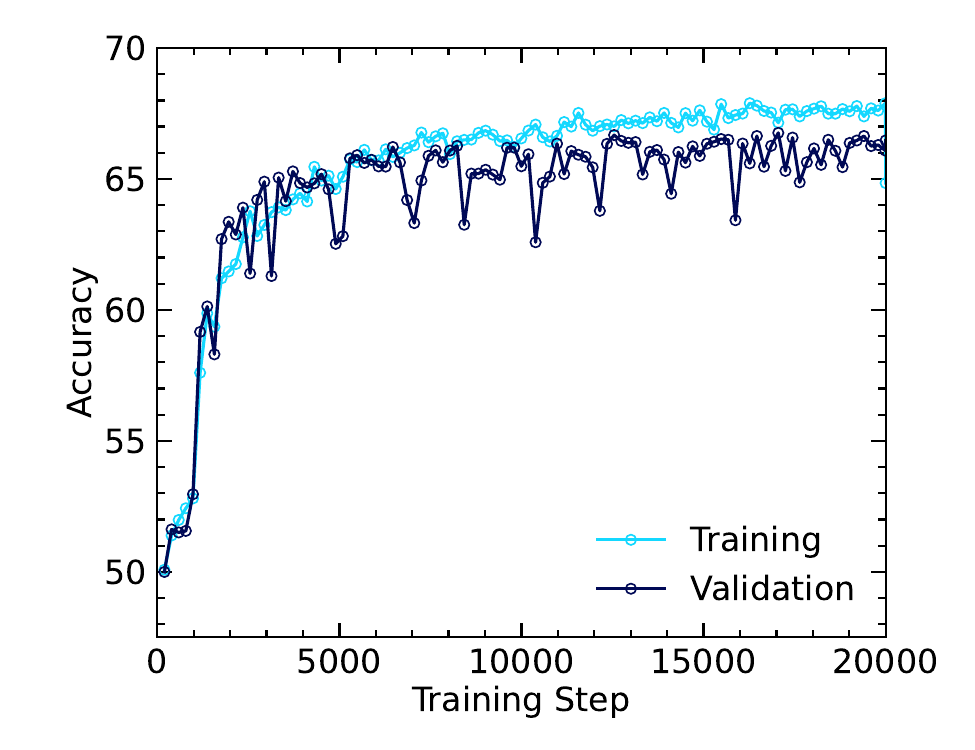}
   \caption{IMDb task.}
\end{minipage}
\hfill
\begin{minipage}{0.3\textwidth}
    \centering
    \includegraphics[width=\textwidth]{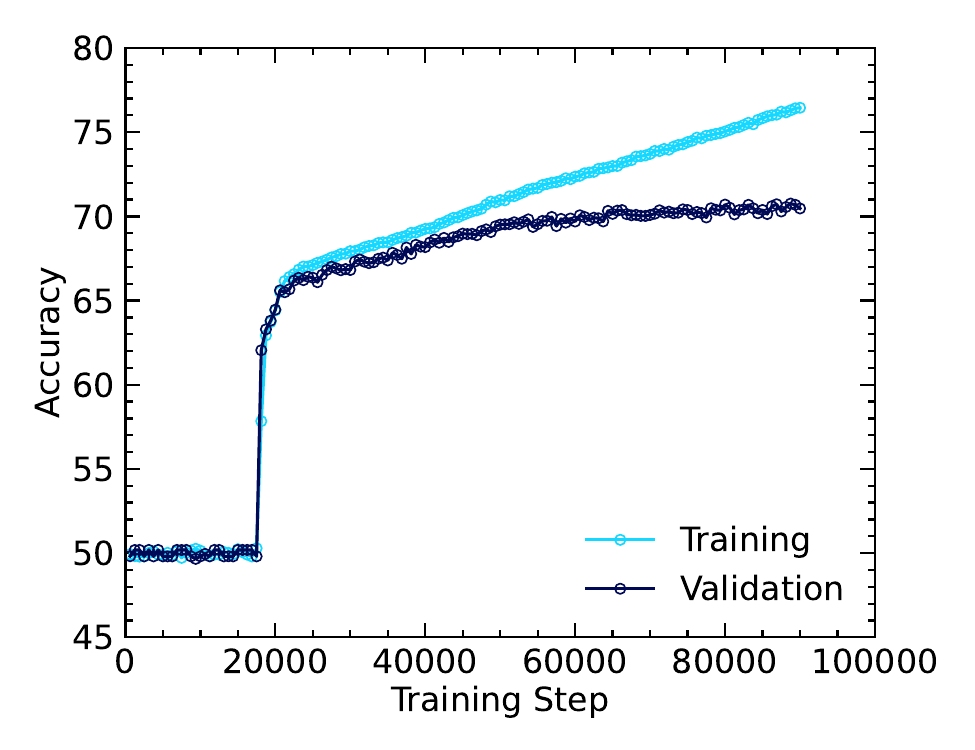}
    \caption{Pathfinder-32 task.}
\end{minipage}
\begin{minipage}{0.3\textwidth}
    \centering
    \includegraphics[width=\textwidth]{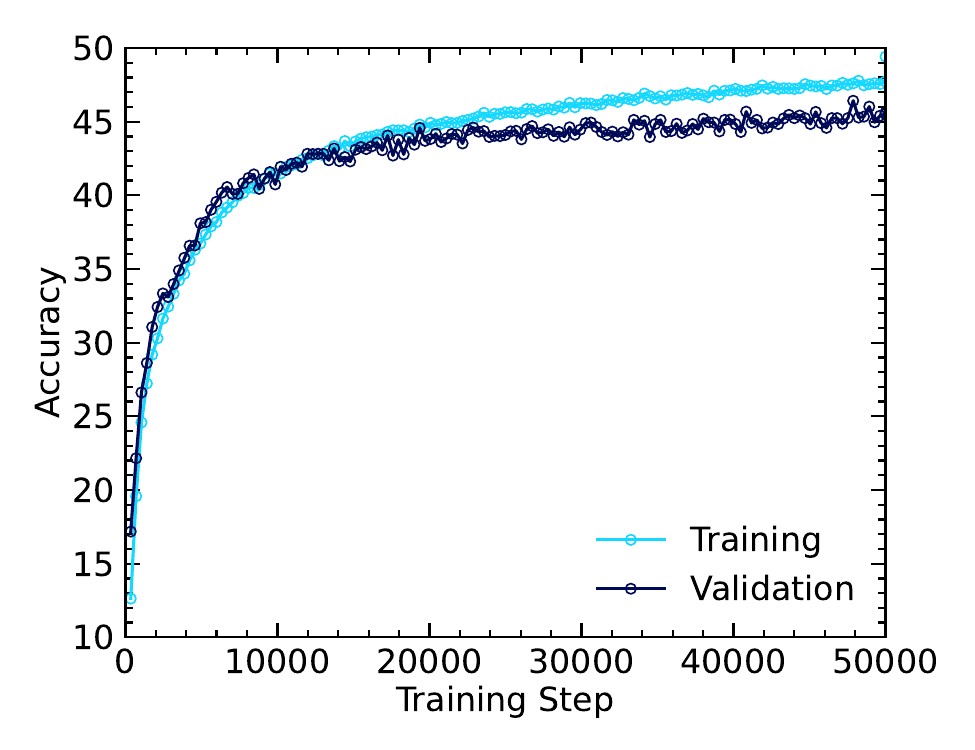}
    \caption{Cifar-10 task.}
\end{minipage}
\hspace{0.4cm}
\begin{minipage}{0.3\textwidth}
    \centering
    \includegraphics[width=\textwidth]{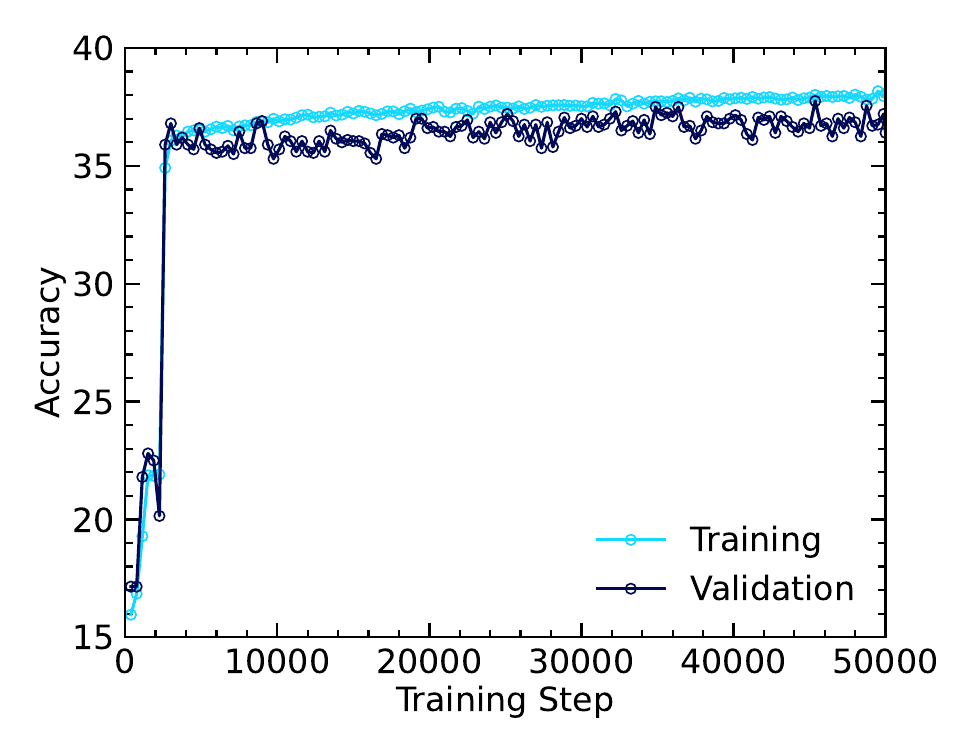}
    \caption{ListOps task.}
\end{minipage}
\caption{Performer standard training accuracy on the different tasks of \gls{lra}.}
\label{fig:fptrainingacc}
\end{figure}

\begin{figure}[h!]
\centering
\begin{minipage}{0.3\textwidth}    
    \includegraphics[width=\textwidth]{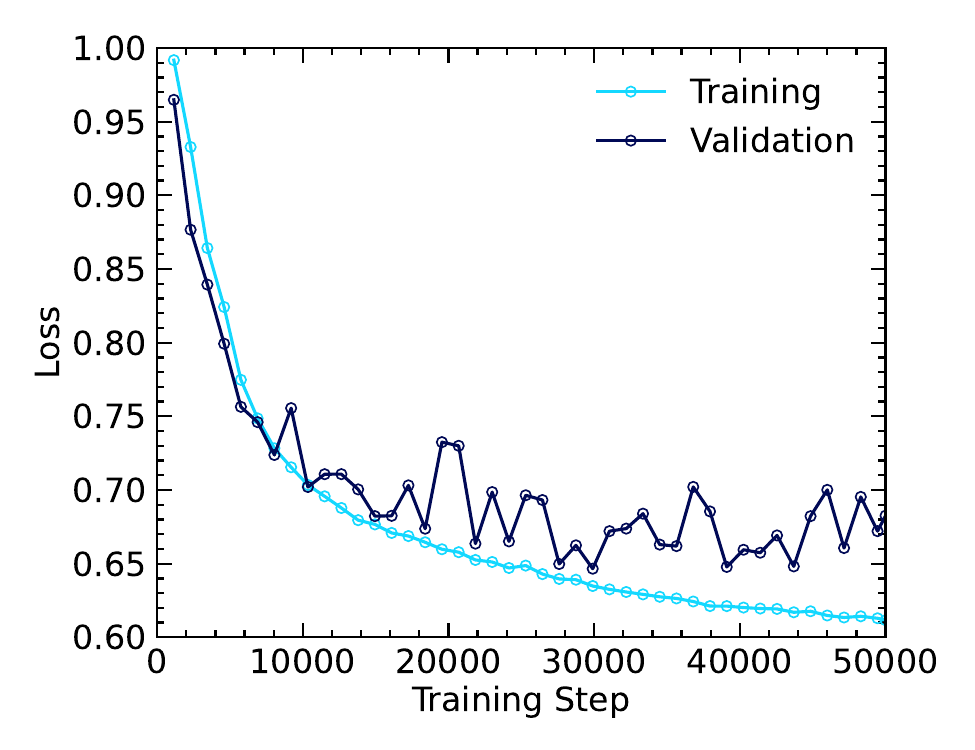}
    \caption{AAN task.}
    \label{fig:fpaan}
\end{minipage}
\hfill
\begin{minipage}{0.3\textwidth}
    \includegraphics[width=\textwidth]{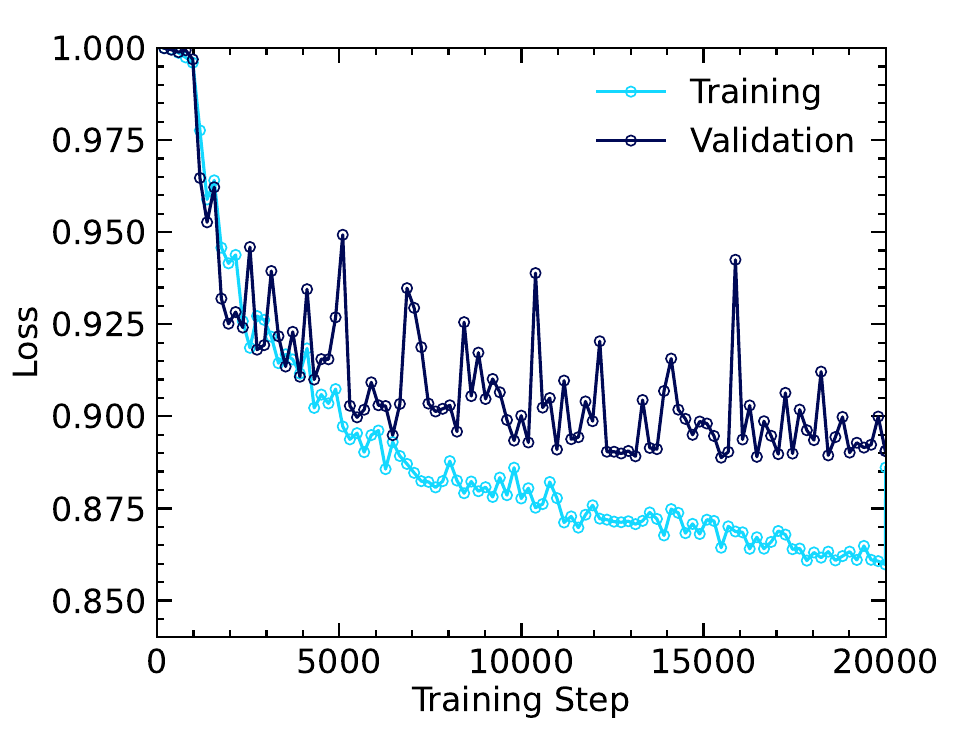}
   \caption{IMDb task.}
\end{minipage}
\hfill
\begin{minipage}{0.3\textwidth}
    \centering
    \includegraphics[width=\textwidth]{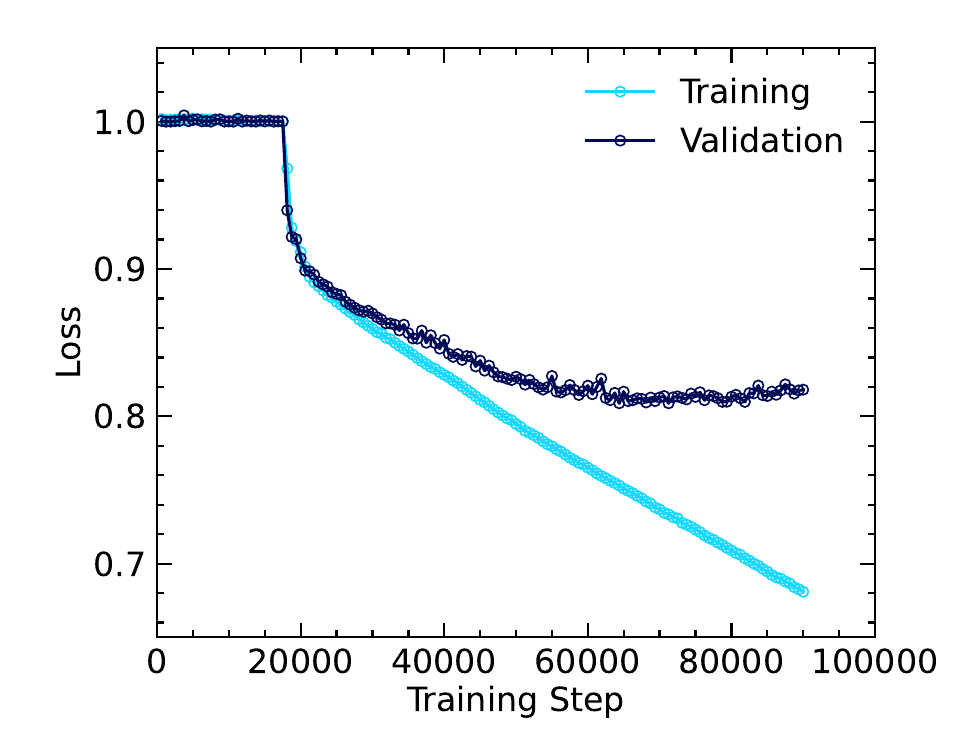}
    \caption{Pathfinder-32 task.}
\end{minipage}
\begin{minipage}{0.3\textwidth}
    \centering
    \includegraphics[width=\textwidth]{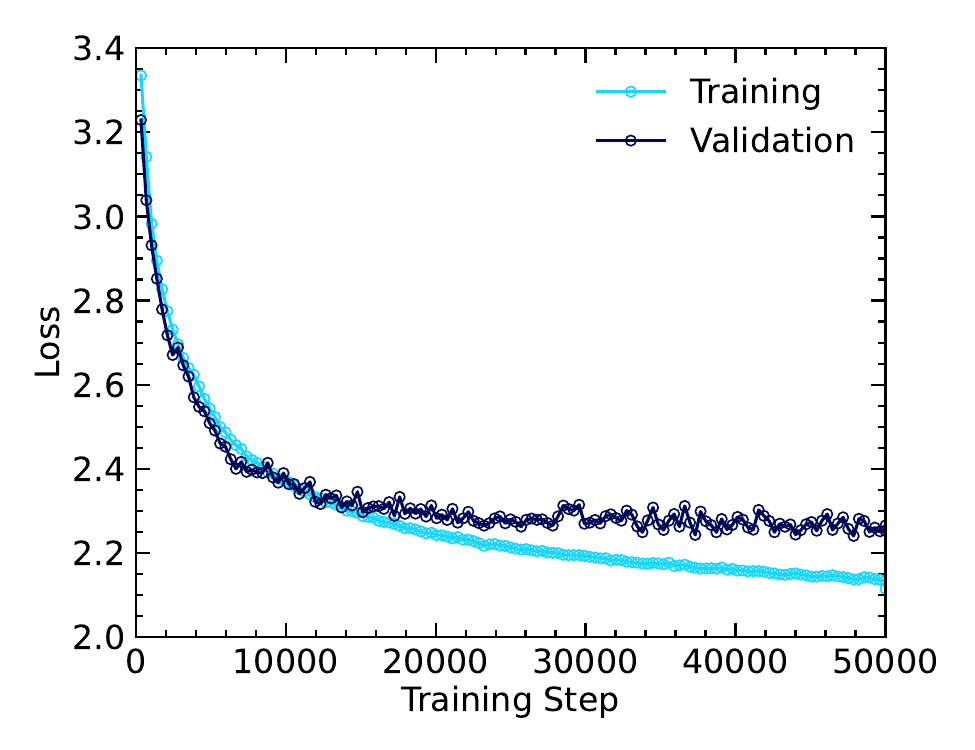}
    \caption{Cifar-10 task.}
\end{minipage}
\hspace{0.4cm}
\begin{minipage}{0.3\textwidth}
    \centering
    \includegraphics[width=\textwidth]{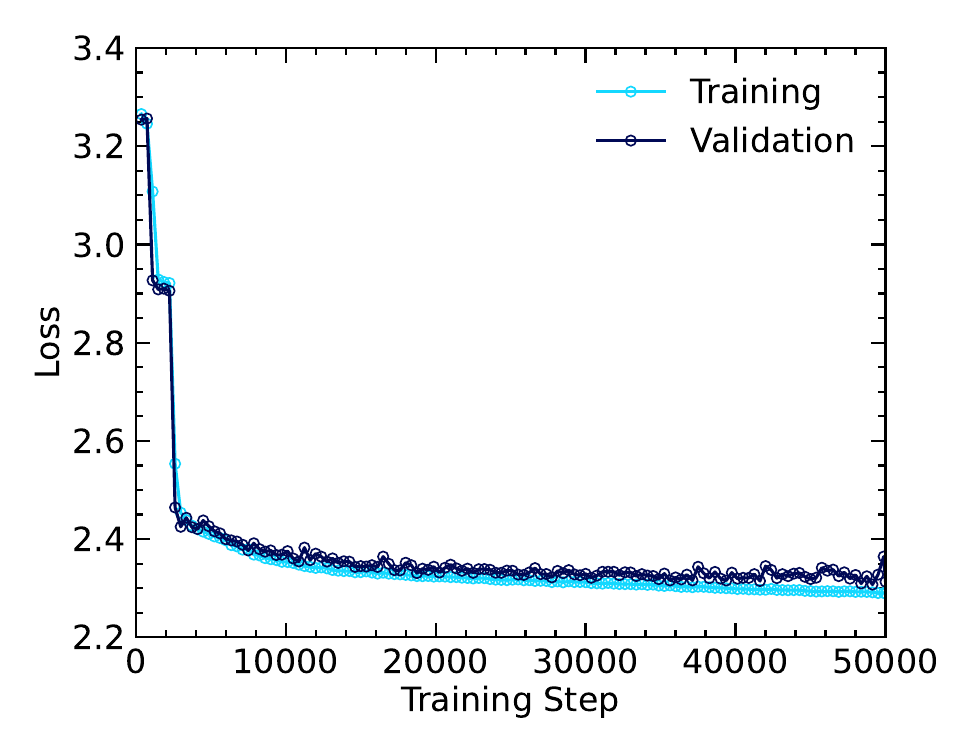}
    \caption{ListOps task.}
\end{minipage}
\caption{Performer standard training loss on the different tasks of \gls{lra}.}
\label{fig:fptrainingloss}
\end{figure}

\clearpage

Generally, the re-training process for the Perfomer allowed us to gain two interesting insights about this model. 
Firstly, we observe that the initialization technique for the embedding layer has a major impact on the convergence of the training for some tasks, especially for one of the tasks: Pathfinder.
For this task, the initial weights of the embedding layer were drawn from a Gaussian distribution with zero mean and unit variance, $\mathbf{w} \sim \mathcal{N}(0,1)$.
However, with this particular initialization the optimization process could not converge to any valid solution during training, and the trained models consistently achieved random chance accuracy across a wide pool of different training and model hyper-parameters.
The problem was only solved by changing the initialization technique back to the standard Transformer embedding layer initialization technique, where the weights of the embedding layer are initialized sampling from a different Gaussian distribution, $\mathbf{w} \sim \mathcal{N}(0, d^{-0.5})$ where $d$ is the embedding dimension.
We speculate that this behaviour can be observed because the model is under-parametrized for the task, and as a result a sub-optimal initialization causes the optimization to get stuck in some local optima, failing to find a meaningful solution for the task.

A second interesting insight that we gain during the re-training of the networks regards the re-sampling of the mapping matrix used in the FAVOR+ mechanism. 
This matrix, whose columns are the random feature vectors drawn from the feature space ($\omega \in \mathbb{R}^d$), was not initially re-drawn periodically during our training of the model.
This caused a huge discrepancy between the accuracy observed during training (and validation, which is performed at the end of each training epoch) and that measured for the test set, illustrated in Supplementary Fig. \ref{fig:resampling} (left). 
The reason behind this phenomenon lies in the implementation of the model: the mapping matrix is not considered a parameter of the model and it is consequently re-initialized every time the model gets loaded into memory.
As a result, the mapping matrix that was used to evaluate the validation samples was the same as the one used during training (no accuracy drop in validation), while the mapping matrix used to evaluate the test samples was different from the one used during training (drop in accuracy for the test partition), as the test evaluation was performed only after the training finished by re-loading the best-performing model in terms of validation accuracy. 
The model was effectively overfitting to the mapping matrix used during training, failing to learn a more general solution to the optimization problem.

Two possible solutions can be implemented to solve this issue. The first one consists of saving the mapping matrix along all the other weights of the model, to be able to use the matrix both at training and inference time. 
On the other hand, another viable solution is to implement the re-sampling directly during training, to avoid overfitting to a specific set of sampled feature vectors and strengthen its generalization capabilities. 
We eventually opted for the latter, as re-sampling could reduce the risk of learning with a poor set of sampled features and, most importantly, increase the robustness of the model to the noise introduced by the \gls{aimc} hardware.
Moreover, the second option is conceptually correct, as the FAVOR+ approximation should not depend on specific sampled features but only on their common distribution from which the vectors are sampled, while the first one would only help the Transformer to learn a ``shortcut'' in the optimization problem.
Supplementary Fig. \ref{fig:resampling} (center) shows the generalization behaviour when the re-sampling is used during training.
This time, the model succeeds to correctly generalize to unseen mapping matrices at test time, and the gap between validation and test scores is (almost) fully closed.

To verify that the model is effectively learning to generalize to the implicit distribution of the features and has not just become invariant to any mapping matrix used in the kernelized attention (by leveraging, for example, the information that still flows through the residual connections in the model) we set up an additional experiment.
We train the model with feature re-sampling and evaluate it during validation with mapping matrices sampled using the correct distribution, but then we evaluate the model on the test partition using a mapping matrix sampled from a different distribution.
In particular, we sampled the matrix from a Poisson distribution with $\lambda=1$.
The results for this experiment are included in Supplementary Fig. \ref{fig:resampling} (right).
We can observe how the model again fails with the mapping matrix sampled from a different distribution, achieving slightly more than chance level accuracy on the test set.
This indeed confirms our hypothesis that the model is effectively adapting to the correct underlying distribution of the sampled features, and is not just becoming invariant to the mapping altogether.

\begin{figure}[h!]
    \centering
    \includegraphics[width=\textwidth]{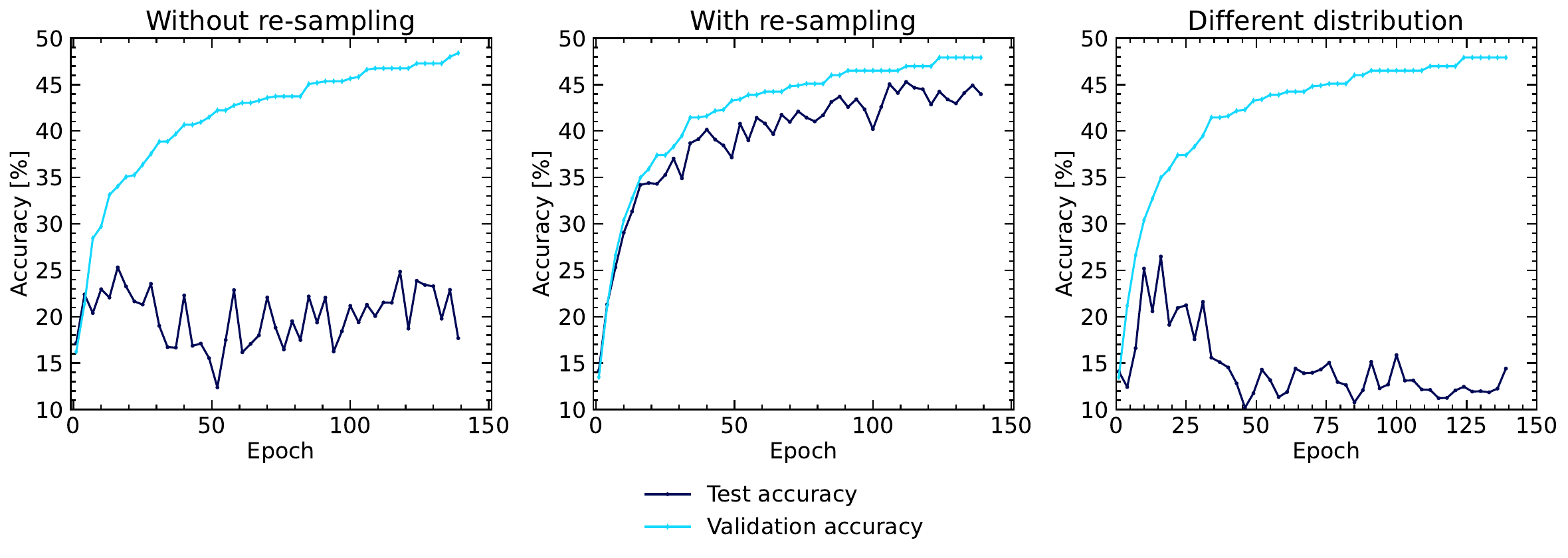}
    \caption{\textbf{Effect of re-sampling the mapping matrix used in FAVOR+ during training on the test accuracy, measured at the end of each epoch on the validation and test partitions of Cifar-10.} While the evaluation on the validation split is performed with the same mapping matrix used during training, the evaluation on the test split is performed with a different mapping matrix. Not using re-sampling during training (left) results in higher validation accuracy, but in a large gap between validation and test accuracy. On the other hand, when re-sampling is used (center), the convergence is slower, but the validation and test accuracies are much closer.
    As a sanity check, we evaluate the model (trained with re-sampling from the right feature distribution) on a mapping matrix sampled from a completely different distribution (right), the Poisson distribution with $\lambda=1$. As expected, the accuracy of the model drops, meaning that re-sampling during training does not make it invariant to \textit{any} mapping matrix used in the approximation, but to matrices sampled from the same distribution.
    The main insight that can be gained from this observation is the following: re-sampling the FAVOR+ mapping matrix during training allows to avoid overfitting to a specific matrix and to learn a more general and robust model.}
    \label{fig:resampling}
\end{figure}

Finally, we report the floating-point precision evaluation results on the full test partition for our standard and hardware-aware trained models for each task of the \gls{lra} benchmark.
For each task, we choose the best model on the basis of the best validation accuracy achieved during training.
The results are included in Supplementary Table \ref{tab:trainingacc}.
We also include previous results from\cite{chen2021skyformer} for the Performer model, and measure the difference with respect to their implementation for the mean accuracy on the benchmark ($\Delta$).

\renewcommand{\arraystretch}{1.5}
\begin{table}[h!]
  \begin{center}
    \begin{tabular}{ p{5cm} | P{1.3cm} P{1.3cm} P{1.3cm} P{1.3cm} P{1.3cm} |P{1.3cm} P{1.3cm} } 
      \toprule
      \textbf{Model} & \textbf{IMDb} & \textbf{ListOps} & \textbf{AAN} & \textbf{Pathfinder}  & \textbf{Cifar-10} & \textbf{Average} & $\bm{\Delta}$ \\ 
      \hline
      Performer\cite{chen2021skyformer} & 64.19 & 38.02 & 80.04 & 66.30 & 41.43 & 58.00 & - \\
      $\text{Performer}^\text{Vanilla training}$ & 66.64 & 37.61 & 77.94  & 70.28  & 45.95  & 59.69 & $+1.69$ \\
      $\text{Performer}^\text{HWA training}$ & 66.46 & 38.63 & 78.07  & 71.61  & 46.17  & 60.19 & $+2.19$ \\
      \bottomrule
    \end{tabular}
      \caption{Performer accuracy comparison on the full test partition between our implementations and previously reported results.}
    \label{tab:trainingacc}
  \end{center}
\end{table}

\clearpage

\section{Replicated Kernel Approximation Baselines}
\label{sup:replication}

Here, the experimental results on kernel approximation techniques (in particular \gls{rff}, \gls{orf}, and \gls{sorf} for the \gls{rbf} and arc-cosine kernels) are replications of the experiments conducted by Liu et al. 2021~\cite{liu2021random}. 
The main goal of this replication study is to validate our framework against previous implementations of the studied kernels and approximation techniques.
Hence, we limit the replication to a single dataset, IJCNN01, out of the six original datasets that are supported by our framework. 
In this experiment, we measure the approximation error, quantified according to the approximation metric defined in the Results section of the main paper on a subset of $1000$ samples of the test set, and the downstream classification accuracy, measured on the predictions of a simple ridge classifier trained on the random features extracted from the samples. 
We evaluate both metrics for $5$ different $\log_2(m/d)$ ratios, where $d$ is the original number of features and $s$ is the number of sampled features; in the dataset used for these experiments, $d = 16$, and hence the number of sampled features is $s = \{32, 64, 128, 256, 512\}$. 
For each combination of kernel/approximation, the results are averaged across $10$ different random seeds, in order to reduce the variance introduced by the random sampling.
Our results are shown in Supplementary Fig. \ref{liureplication}.
In general, they are consistent with the original results from the experiments of the survey paper. It can be observed that, as expected, the two methods that impose an orthogonality constraint on the sampled features (\gls{orf} and \gls{sorf}) perform better than \gls{rff}, especially for lower log ratios. Furthermore, it can be observed that the variance of the estimation, for both metrics, tends to decrease in the number of sampled features, and that often the unstructured method (\gls{rff}) converges to the orthogonal ones for higher numbers of sampled features.

\begin{figure}[h!]
    \centering
    \includegraphics[width=\linewidth]{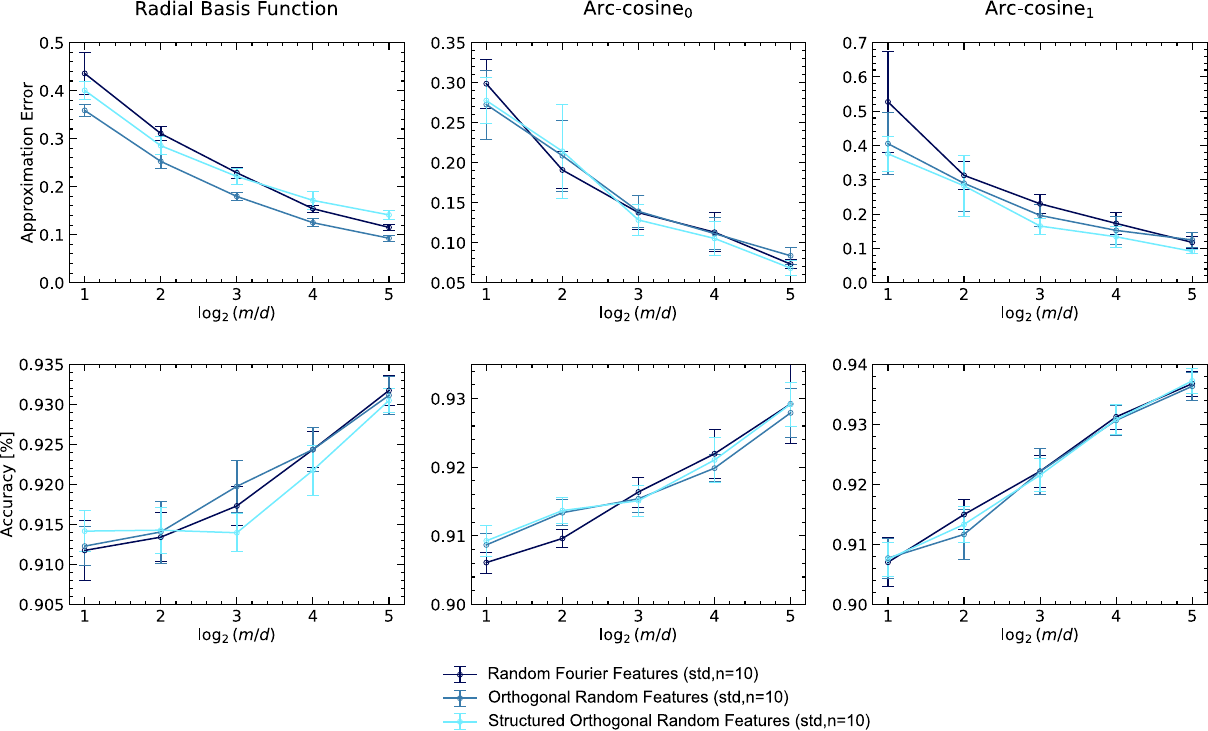}
    \caption{\textbf{Replication experiments for the \gls{rbf} and arc-cosine kernel.} In this figure we replicate the experiments of a previous review paper on kernel approximation techniques from Liu et al.\cite{liu2021random}  on the approximation error and downstream classification accuracy of two kernels (\gls{rbf} and arc-cosine) and three approximation techniques (\gls{rff}, \gls{orf}, and \gls{sorf}). These results are obtained on the IJCNN01 dataset. The approximation error, defined as $||K-\hat{K}||_F/||K||_F$, is reported in the first row, while the downstream classification accuracy is shown in the second row. The obtained results match with the reference paper. The experiments are repeated with $10$ random seeds.}
    \label{liureplication}
\end{figure}

A similar replication experiment was performed to validate the FAVOR+ approximation implemented in the framework, this time trying to match the results obtained by Choromanski et al. 2021\cite{choromanski2020rethinking} in their Figure 4, where the \gls{mse} of the approximation output is compared across different feature implementations (Orthogonal vs IID features and trigonometric sin/cos vs positive features). 
The approximation matrix is computed on $Q$, $K$, and $V$ matrices randomly generated from a normal Gaussian distribution $\mathcal{N}(0, 1)$, with input sequence length $L = 4096$ and original dimensionality $d = 16$.
The experiments are repeated with $15$ different seeds, also in this case to average out the variance introduced by the sampled random features. The results of these experiments are shown in Supplementary Fig. \ref{chorreplication}.
As expected, they closely match the experimental results shown in the original paper. The figure replicates the observation that the positive, exponential-based formulation outperforms the trigonometric formulation of the Softmax kernel in terms of \gls{mse} (see the Performer\cite{choromanski2020rethinking} paper for the definitions). The figure also demonstrates that the orthogonality of the weight vectors further reduces the \gls{mse}.

\begin{figure}[h!]
    \centering
    \includegraphics[width=\linewidth]{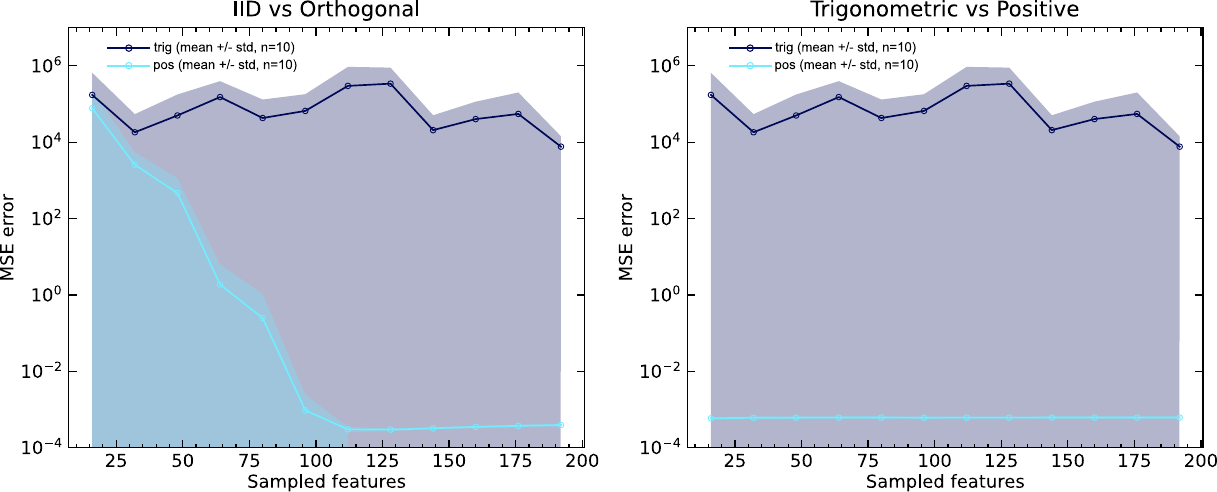}
    \caption{\textbf{Replication experiments for the Softmax kernel.} In this figure we replicate the previous experiments on the approximation error from Choromanski et al.\cite{choromanski2020rethinking}. We measure the approximation error, this time defined as the \gls{mse} between the true and the FAVOR+ approximation of the Softmax kernel matrix. In particular, we compare different type of formulations of FAVOR+, namely using random features vs. orthogonal random features (left) for the trigonometric formulation and using trigonometric vs. positive features (right). Our results are comparable with the reported results\cite{choromanski2020rethinking}. The variances are reported over $10$ different random seeds.}
    \label{chorreplication}
\end{figure}

\clearpage
\section{Comparison to digital von Neumann architectures}
\label{sup:comparison-to-digital}
Supplementary Table \ref{sm:comparison-table} presents the energy consumption and latency for performing kernel approximation on various compute architectures. We compare between our \gls{aimc} hardware~\cite{mt-hermes}, one of the latest CPUs and the NVIDIA A100 GPU~\cite{a100}. Below, we outline the assumptions made for each type of hardware:

\begin{enumerate}
    \item CPU: Intel no longer discloses the number of FLOPs per clock cycle; therefore, we reference the maximum throughput of 1.2288 \gls{tops} at 253W for the Intel 14th Gen i9-14900KF CPU.
    \item GPU: The reported metrics are based on 312 \gls{tops} (assuming FP-16) and 624 \gls{tops} (assuming int8) at a power consumption of 400W. We assume that for the matrix sizes used, data fetching times are subsumed within the compute latency.
    \item \gls{aimc}: Our hardware demonstrates an MVM throughput of 63.1 \gls{tops} while consuming approximately 6.5W.
\end{enumerate}
Supplementary Table \ref{sm:comparison-table} details the latency and energy consumption for various configurations of sequence length $L$, model dimension $d$, and mapping dimension $m$. We omit post-processing operations and focus solely on the mapping, which constitutes the majority of the computation.

\begin{table}[htbp]
\centering
\caption{Comparison of kernel-approximation latency and energy consumption between \gls{aimc} hardware and digital von Neumann based architectures.}
\label{sm:comparison-table}
\begin{tabular}{cccccc}
\toprule
& & AIMC & GPU INT8 & GPU FP16 & CPU \\
\midrule
\multicolumn{2}{c}{\textbf{L = 1024, d = 512, m = 1024}} \\
Latency (ms) & & 0.0170 & 0.0017 & 0.0034 & 0.8738 \\
Energy (mJ)  & & 0.1100 & 0.6883 & 1.3766 & 221.0748 \\
\midrule
\multicolumn{2}{c}{\textbf{L = 1024, d = 1024, m = 2048}} \\
Latency (ms) & & 0.0681 & 0.0069 & 0.0138 & 3.4953 \\
Energy (mJ)  & & 0.4401 & 2.7532 & 5.5064 & 884.2991 \\
\bottomrule
\end{tabular}
\end{table}

As shown in the data, the GPU exhibits an \gls{mvm} throughput that is approximately 9.9 times higher than that of the IBM HERMES Project Chip. However, these calculations assume 100\% utilization of the GPU's Tensor Cores, which is unrealistic at these matrix dimensions. The same holds for the IBM HERMES Project Chip: the above matrices consume only 8 and 32 cores on the IBM HERMES Project Chip, reducing the \gls{tops} to 7.8875 and 31.55, respectively. However, one can utilize the full 63.1 \gls{tops} of the IBM HERMES Project Chip by replicating the weights 8, and 2 times, respectively. One can even further boost the throughput by using multiple instances of the IBM HERMES Project Chip, with each one having a footprint of $12 mm^2$ compared to the $826 mm^2$ of the NVIDIA A100. Furthermore, the NVIDIA A100 was produced using TSMC's N7 process while the IBM HERMES Project Chip was fabricated in 14-nm CMOS.
Besides the throughput at small to moderate problem sizes, the GPU has another fundamental limitation: the power consumption. At peak throughput, the NVIDIA A100 consumes 400 W of power, of which ~70 W are static, compared to the 6.5 W of the IBM HERMES Project Chip. This not only makes the IBM HERMES Project Chip a viable option for deployment at the edge, but also leads to an energy efficiency advantage of between 6.2 to 12.4 times compared to the NVIDIA A100.
Finally, the CPU, while consuming less power than the GPU, offers limited parallelism with only 8 performance cores, achieving a throughput of 1.22 \gls{tops}, which is documented in the table.

\end{document}